\RequirePackage[svgnames]{xcolor}

\documentclass[11pt,letterpaper]{mystyle}

\usepackage[all]{hypcap}
\usepackage[svgnames]{xcolor}
\usepackage[comma,authoryear,compress]{natbib}
\usepackage{hyperref}[citecolor=lightblue]

\hypersetup{
    colorlinks = true,
    citecolor = {YaleBlue},
}

\usepackage{algorithm}
\usepackage{algorithmicx}
\usepackage{algpseudocode}
\usepackage{microtype}
\usepackage{graphicx}
\expandafter\def\csname ver@subfig.sty\endcsname{}
\usepackage{booktabs} %
\usepackage{float}
\usepackage{bigstrut}

\usepackage{amsmath}
\usepackage{amssymb}
\usepackage{mathtools}
\usepackage{amsthm}
\usepackage{mathrsfs}
\usepackage{nicefrac}
\usepackage{dsfont}
\usepackage{enumitem}
\usepackage{subcaption}
\usepackage{graphicx,subfig}
\usepackage{cleveref}
\usepackage{bxcoloremoji}

\usepackage{float}

\usepackage[utf8]{inputenc} %
\usepackage[T1]{fontenc}    %
\usepackage{hyperref}       %
\usepackage{url}            %
\usepackage{booktabs}       %
\usepackage{amsfonts}       %
\usepackage{nicefrac}       %
\usepackage{microtype}      %
\usepackage{graphicx}
\usepackage{subcaption} %
\usepackage{wrapfig}
\usepackage{lipsum}
\usepackage{enumitem}
\usepackage{stackengine}
\usepackage[font=small,labelfont=bf]{caption}
\usepackage{color}
\usepackage{adjustbox}

\usepackage{rotating}
\usepackage{makecell}

\usepackage{colortbl}
\usepackage{multirow}
\usepackage{xspace}
\usepackage{bm}
\usepackage{longtable}
\usepackage{threeparttable}
\usepackage{placeins}
\addtocontents{toc}{\protect\setcounter{tocdepth}{-1}}
\definecolor{ProjectPink}{HTML}{E6007E}

\DeclareMathOperator*{\argmin}{arg\,min}

\tcbuselibrary{skins,breakable,listings}   %

\lstdefinestyle{promptstyle}{
  basicstyle=\ttfamily\scriptsize,
  breaklines=true, breakatwhitespace=false, columns=fullflexible,
  keepspaces=true, showstringspaces=false, tabsize=2,
  postbreak=\mbox{$\hookrightarrow$\space},
  literate=
    {→}{{$\rightarrow$}}1 {←}{{$\leftarrow$}}1
    {—}{{---}}1 {–}{{--}}1
    {≥}{{$\geq$}}1 {≤}{{$\leq$}}1
    {’}{{'}}1 {‘}{{`}}1 {“}{{``}}1 {”}{{''}}1
    {˜}{{\textasciitilde}}1 {°}{{$^\circ$}}1 {×}{{$\times$}}1,
}

\lstdefinelanguage{json}{
  morestring=[b]",
  stringstyle=\color{blue!60!black},
  literate=
    {→}{{$\rightarrow$}}1 {—}{{---}}1 {–}{{--}}1,
}

\newtcblisting[auto counter]{promptbox}[2][]{
  listing engine=listings,
  listing options={style=promptstyle},
  listing only, breakable,
  colback=gray!4, colframe=gray!60, boxrule=0.4pt, arc=2pt,
  left=4pt, right=4pt, top=2pt, bottom=2pt,
  title={\footnotesize\bfseries Listing~\thetcbcounter: #2},
  coltitle=black, colbacktitle=gray!15,
  #1
}

\newtcblisting[use counter from=promptbox]{jsonbox}[2][]{
  listing engine=listings,
  listing options={style=promptstyle, language=json},
  listing only, breakable,
  colback=blue!2, colframe=blue!40!gray, boxrule=0.4pt, arc=2pt,
  left=4pt, right=4pt, top=2pt, bottom=2pt,
  title={\footnotesize\bfseries Listing~\thetcbcounter: #2},
  coltitle=black, colbacktitle=blue!8,
  #1
}

\newcommand{\x}{\mathbf{x}}

\definecolor{blanchedalmond}{rgb}{1.0, 0.92, 0.8}
\definecolor{carmine}{rgb}{0.59, 0.0, 0.09}
\definecolor{lightblue}{rgb}{0.22,0.45,0.70}%

\renewcommand{\mathbf}{\boldsymbol}

\makeatletter
\def\Ddots{\mathinner{\mkern1mu\raise\p@
\vbox{\kern7\p@\hbox{.}}\mkern2mu
\raise4\p@\hbox{.}\mkern2mu\raise7\p@\hbox{.}\mkern1mu}}
\makeatother

\definecolor{amaranth}{rgb}{0.9, 0.17, 0.31}
\definecolor{antiquebrass}{rgb}{0.8, 0.58, 0.46}
\definecolor{antiquefuchsia}{rgb}{0.57, 0.36, 0.51}
\definecolor{chromeyellow}{rgb}{0.31, 0.47, 0.26}

\newtcolorbox{AIbox}[2][]{aibox,title=#2,#1}
\definecolor{lightblue}{rgb}{0.22,0.45,0.70}%
\definecolor{Gray}{gray}{0.95}
\definecolor{Cornsilk}{rgb}{1.0, 0.97, 0.86}

\newcommand{\algonamebf}{\textbf{Dex-One2Many}\xspace}
\newcommand{\algoname}{Dex-One2Many\xspace}

\usepackage[all]{hypcap}

\title{\algoname: Learning Dexterous Manipulation from a Single Human Demonstration}

\runningtitle{\algoname: Learning Dexterous Manipulation from a Single Human Demonstration}

\author{
  Jusuk Lee\textsuperscript{1*},
  Sungha Kim\textsuperscript{1*},
  Yeonsoo Park\textsuperscript{1*},
  Jonguk Cheon\textsuperscript{1},
  Yoonkyo Jung\textsuperscript{2},
  Yongjun You\textsuperscript{1},
  H.~Jin~Kim\textsuperscript{1},
  Jia-Bin Huang\textsuperscript{2},
  Furong Huang\textsuperscript{2,3},
  Youngseok Jang\textsuperscript{4\textdagger},
  Seungjae Lee\textsuperscript{2\textdagger}\\
  {\normalfont\fontsize{10}{12}\selectfont\textsuperscript{1}Seoul National University\quad\textsuperscript{2}University of Maryland, College Park}\quad
  \textsuperscript{3}All Purpose AI\quad
  \textsuperscript{4}KAIST\\
  {\normalfont\fontsize{10}{12}\selectfont
    \href{https://dex-one2many.github.io/}{
      \textcolor{TinaCrimson}{\textbf{https://dex-one2many.github.io}}
    }
  }
}

\authornote{{\fontsize{9}{11}\selectfont\textsuperscript{*}Equal contribution\quad\textsuperscript{\textdagger}Equal advising}}
\correspondingauthor{}

\begin{document}
\hbadness=10000
\addtocontents{toc}{\protect\setcounter{tocdepth}{-1}}
\begin{abstract}
While learning dexterous manipulation from a single human video offers a promising alternative to costly robot demonstrations, many recent methods predominantly imitate demonstrated motions. Such strict motion matching often limits generalization to initial object poses, goal poses, and grasps not shown in the video. Alternatively, discovering a policy via reinforcement learning (RL) allows for broad generalization, but without prior guidance, it struggles with high-dimensional exploration in complex, multi-stage tasks. To address these coupled generalization and exploration challenges, we present \algonamebf, a real-to-sim-to-real framework that learns a generalizable dexterous manipulation policy from a single human video. 
Our key insight is to abstract the video into sequential scene graphs that guide RL, enabling efficient exploration while preserving broad generalizability. The graphs serve as generative constraints for sampling diverse reset states and provide dense rewards for each stage. Because the graphs constrain relations rather than exact poses, these reset states cover object poses and grasps beyond the video, while initializing each stage from them with dense rewards keeps exploration short and guided.
Trained entirely in simulation, \algoname transfers zero-shot to a real multi-fingered hand. Across five tool-use and manipulation tasks, \algoname exceeds baselines by 6.5\% in seen configurations, while its robust generalization widens this gap to 71\% in unseen scenarios.
\end{abstract}

\maketitle
\vspace{3mm}

\begin{figure*}[!htbp]
    \centering
    \vspace{-0.5em}
    \includegraphics[width=1\textwidth]{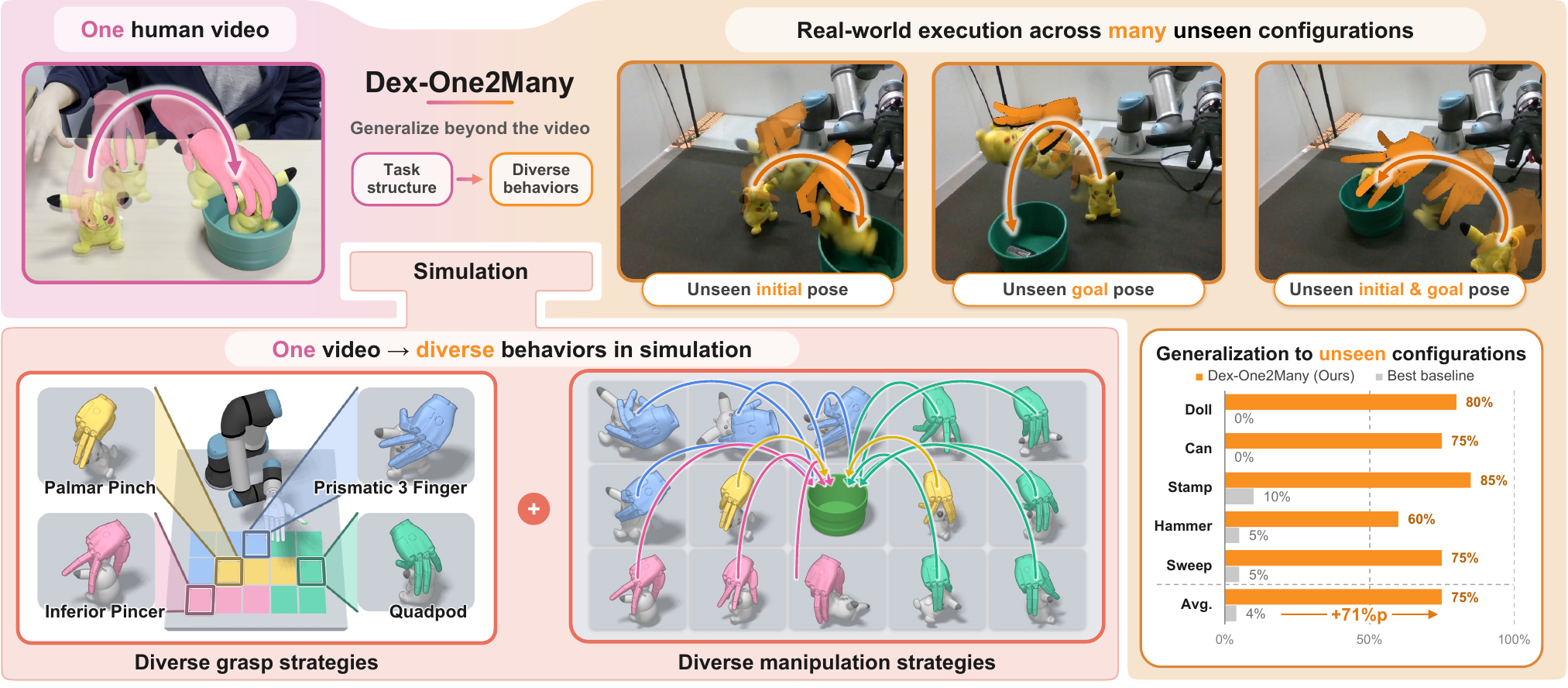}
    \vspace{-1.5em}
    \caption{\textbf{From one human video to many dexterous behaviors.} Given \emph{one} human video of a manipulation task, \algoname learns a dexterous policy entirely in simulation and deploys it zero-shot in the real world. By training across diverse simulated task configurations, the resulting policy executes the task across \emph{many} unseen configurations beyond the execution shown in the video.
    }
    \label{fig:thumbnail}
    \vspace{-1em}
\end{figure*}

\section{Introduction}
\label{sec:1}
Dexterous manipulation is a promising direction for enabling robots to perform complex object interactions, but collecting enough high-quality robot demonstrations remains a major bottleneck. For high-degree-of-freedom hands, teleoperation interfaces are particularly costly to design and operate, and the collected motions are tied to a specific robot morphology. Human videos offer a way around this bottleneck: they are easy to collect without a robot or a teleoperation rig and contain rich visual and kinematic information about hand--object interactions. Recent methods exploit this to learn dexterous manipulation from just a single human video~\citep{human2sim2robot, doasido, video2sim2real, chord}.

However, these single-video methods mostly convert the video into execution-level guidance, such as reconstructed object trajectories~\citep{human2sim2robot}, retargeted hand motions~\citep{doasido, video2sim2real}, or contact cues~\citep{chord}. Such guidance facilitates high-dimensional dexterous policy learning, but also ties the resulting policy to the motion and spatial configuration shown in the video. Even for the same task, these methods generalize poorly to initial object poses, goal poses, and grasps not shown in the video.

Learning a dexterous policy that generalizes beyond the execution shown in the video raises two tightly coupled challenges. The first is \emph{task abstraction}: deciding what to preserve from the video and what to leave free to vary. Preserving too much reduces the policy to replaying the recorded motion; preserving too little leaves the task underspecified. The second is a \emph{hard exploration problem}: the more that is left free to vary, the more of the state space the policy must explore. Without execution-level guidance, reinforcement learning (RL) struggles to discover multi-finger, contact-rich interactions in this high-dimensional space, especially in multi-stage tasks.

\begin{figure}[t]
  \centering
  \includegraphics[width=\linewidth]{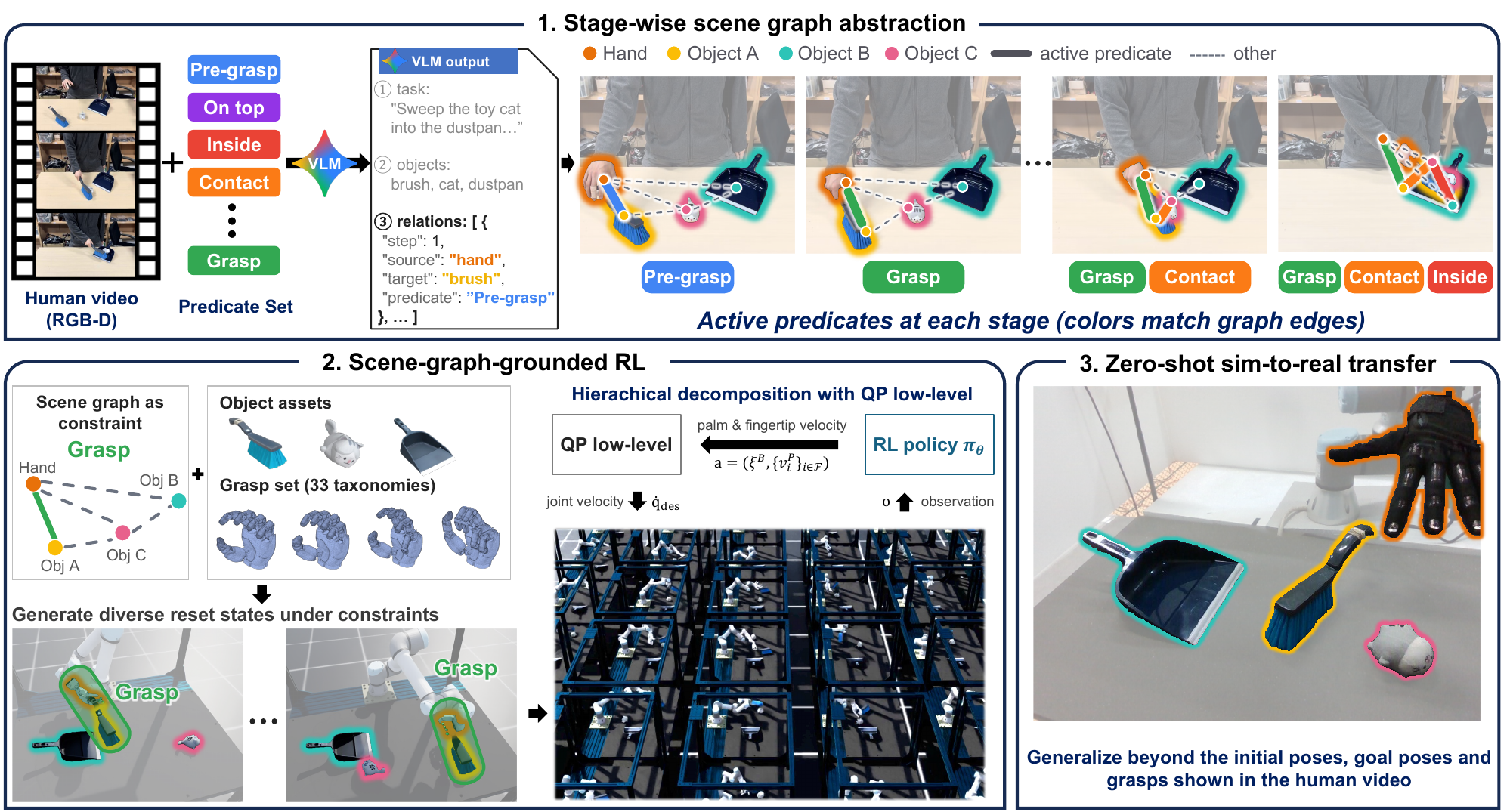}
  \caption{\textbf{Overview of \algoname.} (1) From a single human video, a VLM extracts a stage-wise scene graph of object relations.  (2) Each stage graph constrains the sampling of object poses and synthesized grasps, yielding verified reset states for every stage. An RL policy trained from these resets with predicate-derived rewards acts in task space through a QP controller. (3) The policy transfers zero-shot to unseen initial poses, goal poses, and grasps on real hardware.
  }
  \label{fig:overview}
  \vspace{-10pt}
\end{figure}

To address both challenges, we present \algonamebf, a real-to-sim-to-real framework that learns a generalizable dexterous manipulation policy from a single human video (Fig.~\ref{fig:thumbnail}). Our key idea is to extract not the video's motion but its task structure, a stage-wise scene graph specifying which hand--object and object--object relations must hold, and in what order, while leaving RL to learn how the hand acts. To this end, we propose two techniques (Fig.~\ref{fig:overview}).

First, \emph{Stage-Wise Scene Graph Abstraction} reduces the video to the relations the task requires, such as the hand holding an object or an object resting inside a container, and to the order in which they occur. Object poses, goal poses, and grasps remain free to vary. The abstraction splits the video into stages when these relations change and encodes each stage as a graph of predicates over the hand, objects, and their functional parts. Each stage graph describes not one recorded state but a set of states that all preserve the task.

Second, \emph{Scene-Graph-Grounded RL} uses each stage graph as a generative constraint: it samples reset states that satisfy the stage's relations, with diverse multi-finger grasps, and keeps only those that pass our simulation checks. One video thus becomes many training states across the whole workspace. The accepted states form the initial-state distribution of an MDP whose observations and rewards are derived from the graph. Starting episodes at every stage eases exploration: the policy can practice the last stage before mastering the first, rather than reaching it only by chance. Within each stage, predicate-derived rewards give a dense signal of progress toward satisfying the stage's relations, without reference to a demonstrated trajectory. The policy therefore learns the task structure rather than the recorded execution.

Trained entirely in simulation, the resulting policy completes the demonstrated task zero-shot on a real multi-fingered hand across initial and goal object poses not shown in the video. We evaluate five tasks spanning direct object manipulation and tool use on a UR3 arm with a 20-DoF Wuji~1 hand. On configurations with initial and goal poses not shown in the video, \algoname achieves a 60--85\% real-world success rate, while the baselines remain at or below 15\%. Our main contributions are as follows:
\begin{itemize}[leftmargin=1.2em, itemsep=0pt, topsep=2pt]
    \item \textbf{Stage-Wise Scene Graph Abstraction:} We represent a human demonstration as a stage-wise scene graph that retains task-defining structure---the required hand--object and object--object relations and their ordering---while discarding execution-specific motion.
    
    \item \textbf{Scene-Graph-Grounded RL:} We use each stage's scene graph to generate diverse reset states through the task, enabling the policy to explore alternative grasps and motions while grounding its observations and rewards from the graph.

    \item \textbf{Zero-shot sim-to-real transfer on unseen configurations:} Across five real-world tasks, policies learned from a single human video achieve 75\% mean success on configurations not shown in the input video, exceeding the next-best baseline by 71 percentage points.
\end{itemize}

\section{Related Work}
\label{sec:2}
Generalizing dexterous policies beyond the execution shown in a single human video raises two key challenges identified in the introduction: \emph{task abstraction} and \emph{hard exploration}. We review the closest work in these two directions, along with broader related work, in Appendix~\ref{app: related_work}.

\noindent\textbf{Task Structure Extraction from Human Videos.}
Human videos have been used to learn reusable visual representations~\citep{r3m,dynaflip}, rewards~\citep{vip, graphirl, virl}, point tracks~\citep{track2act}, latent plans~\citep{mimicplay}, or action priors~\citep{shaw2023videodex} from large video collections. In the single-video setting, however, generalization depends largely on how the video is abstracted. Prior work abstracts the video into keypoints, keyframes, object-centric graphs, or object pose trajectories, then adapts the demonstrated execution to a new scene by warping or tracking it, or by synthesizing imitation data around it~\citep{kvil, graphvisualimitation, ditto, orion, okami, yoto, mimicfunc, real2gen}. Because what transfers is the demonstrated motion, adaptation extends only as far as that motion can be reshaped. Relational task specifications provide a more abstract alternative and are well established in symbolic and scene-graph planning~\citep{fikes1971strips, sayplan}, and vision-language models have extracted symbolic task plans from a human video~\citep{wake2024gpt} and stage-wise relational constraints from language~\citep{rekep}. Building on this relational view, our \emph{Stage-Wise Scene Graph Abstraction} extracts task structure from the single video, while discarding execution details, encoding each stage as a graph whose edges are predicates stating which hand--object and object--object relations must hold, in the order the video shows. Unlike ORION's graphs~\citep{orion}, our stage-wise scene graph carries no coordinates or keypoint trajectories from the video, so each graph describes not one state but the set of all states that satisfy its predicates. Rather than a plan to execute, a reference to adapt, or a cost to optimize, each graph serves as a constraint that sampled training states must satisfy, leaving object poses, goal poses, and grasps free to vary.
RL then learns how the hand acts instead of inheriting the demonstrated motion.

\noindent\textbf{Learning Dexterous Manipulation from a Single Human Video.}
Dexterous manipulation poses a hard exploration problem for RL because multi-fingered hands must coordinate many joints through contact-rich interactions~\citep{rajeswaran2017learning}. Prior methods that learn dexterous manipulation from human video therefore extract execution-level guidance, such as object pose trajectories used as a tracking reward~\citep{human2sim2robot, chord, sharma2026one, chen2025vividex, dexmachina}, retargeted hand motions used as references to imitate or refine~\citep{qin2022dexmv, maniptrans, spider, doasido, video2sim2real, hudor}, or demonstrated contacts~\citep{chord, sharma2026one}. This guidance makes exploration tractable but ties the resulting behavior to the demonstrated motion, even when another initial object pose, goal pose, or grasp requires a different execution of the same task. As a result, their generalization is largely limited to local variations around the demonstrated configuration~\citep{human2sim2robot, video2sim2real,chord}.\footnote{\scriptsize Human2Sim2Robot: initial position within 4\,cm$\times$4\,cm, fixed goal; Video2Sim2Real: initial/goal within $\pm$5\,cm, $\pm10^\circ$.} Removing execution-level guidance, however, creates the second challenge: how to make RL explore efficiently from task structure alone. Reset-based RL alleviates hard exploration by starting episodes near a known goal~\citep{florensa} or along a demonstration~\citep{backplay, montezuma, rfcl, demostart}, so the policy learns from later states before it can reach them. Demonstration-based resets, however, inherit the demonstrated execution. More recent approaches such as OmniReset~\citep{omnireset} and Video2Policy~\citep{video2policy} reduce this dependence by generating reset states or task programs, but rely on user-specified task information or video-derived object states and have only been demonstrated with parallel-jaw grippers. Our approach grounds resets, observations, and rewards in a stage-wise scene graph derived from the human video, so it derives the task structure without imitating the execution.

\section{Method}
Given a single demonstration of a manipulation task, our goal is to learn a dexterous policy that generalizes to diverse spatial configurations, including randomized object poses and novel grasps, beyond the original video. To achieve this, we first abstract the demonstration into a stage-wise scene graph (Sec.~\ref{sec:scene_graph_abstraction}). 
We then formulate policy learning as a Markov decision process (MDP) grounded in this graph, with observations, rewards, and an initial-state distribution constructed from it without task-specific design (Sec.~\ref{sec:rl}). The learned policy then transfers zero-shot to the real robot (Sec.~\ref{sec:sim2real}). Fig. \ref{fig:overview} summarizes the full pipeline.

\subsection{Task Abstraction as a Stage-wise Scene Graph}
\label{sec:scene_graph_abstraction}

We abstract the human video as a sequence of stage-wise scene graphs that preserve which relations must hold and in what order, while allowing object poses, goal poses, grasps, and trajectories to vary. Each graph represents one stage of the task: its nodes identify the hand, task-relevant objects, and their functional parts, while its edges describe the relations characterizing that stage. When the same task is performed from different configurations, the poses and trajectories may differ, but the required relations and their order remain the same. We refer to this abstraction as \emph{Stage-Wise Scene Graph Abstraction}.

To encode the relations carried by the scene-graph edges without fixing the poses observed in the video, we use a predefined, task-independent \emph{predicate} vocabulary (Table~\ref{tab:predicates}). Some predicates specify relations that must hold, such as \texttt{grasp}, \texttt{contact}, and \texttt{inside}, while others describe proximity toward the next relation to be established, such as \texttt{pre\_grasp} and \texttt{pre\_inside}. The former describe what must hold in the current stage, whereas the latter indicate which relation should be established next. Together, these predicates describe task progression without fixing the specific configuration shown in the video.

We recover these stage-wise graphs from the video using a vision-language model (VLM) that abstracts the task directly. Given the video and the predicate vocabulary, the VLM identifies task-relevant entities and functional parts as graph nodes and infers the predicate relations between them over time. We introduce a new stage whenever the predicate relations change, either because a new relation is added or because a relation being approached becomes established, such as \texttt{pre\_grasp} transitioning to \texttt{grasp}. This yields an ordered sequence $\mathcal G_{1:K}=(\mathcal G_1,\ldots,\mathcal G_K)$. Fig.~\ref{fig:overview} illustrates how the stage-wise scene graph captures task progression in the sweeping task: a new stage is introduced when \texttt{pre\_grasp} becomes \texttt{grasp}; in later stages, the established \texttt{grasp} and \texttt{contact} relations remain as constraints, while the toy cat is moved toward the dustpan until the new \texttt{inside} relation is established.

\begin{figure}[t]
  \centering
  \vspace{-5pt}
  \includegraphics[
    width=\linewidth,
    trim=0mm 5mm 0mm 0mm
  ]{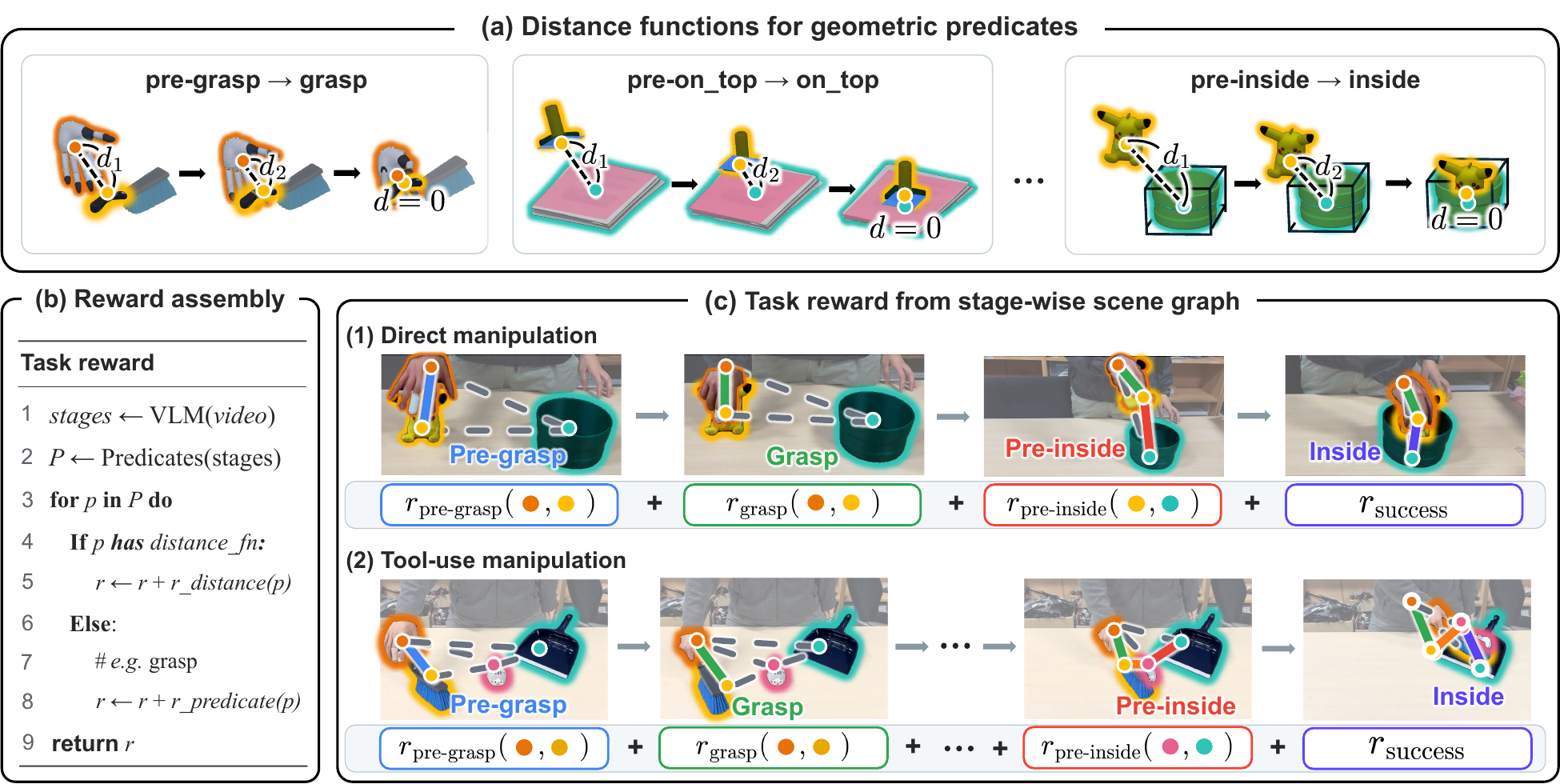}
  \caption{
    \textbf{Using scene-graph predicates for policy learning.}
    (a) Predicates that can be expressed geometrically are mapped to Cartesian-space distance functions.
    (b) The predicates in a stage-wise scene graph determine which reward terms are assembled into the task reward.
    (c) The same assembly rule is instantiated for direct and tool-use manipulation from their respective scene graphs.
}
  \label{fig:reward_overview}
  \vspace{-12pt}
\end{figure}

\subsection{Scene-graph-grounded Reinforcement Learning}
\label{sec:rl}
Given the scene-graph sequence $\mathcal G_{1:K}$ from Sec.~\ref{sec:scene_graph_abstraction}, we formulate policy learning as a finite-horizon MDP $\mathcal M=\langle\mathcal S,\mathcal A,p,r,\gamma,\rho_0,T\rangle$. Here, $\gamma$ and $T$ denote the discount factor and episode horizon; the simulator provides transition dynamics $p$, and $\mathcal A$ consists of desired palm and fingertip velocities tracked by a low-level QP controller (Sec.~\ref{sec:sim2real}). The scene graph defines the policy observation $o$ from $\mathcal S$, the reward $r$, and the initial-state distribution $\rho_0$, as described below.

\textbf{Observation from the scene graph.}
We construct the policy observation $o$ from the task-relevant state associated with the scene graph's nodes and edges.
Nodes are represented by role: the hand node provides robot proprioception, object nodes encode object geometry and 6D pose, and functional-part nodes localize regions such as the grasp region. Edges capture the relative 6D pose between connected entities. Because these encodings are shared across tasks, the same observation construction applies to different task graphs. Concatenating these node and edge features forms the policy observation, making the observation design reusable (see Appendix~\ref{app:observations} for details).

\textbf{Task reward from the scene graph.}
Each scene-graph edge encodes a predicate between two entities, which we map to a geometric or physical reward. Geometric predicates use the distance metric in Table~\ref{tab:predicates} as a dense progress signal, a 3D Cartesian distance for spatial relations (e.g., to a grasp region for \texttt{pre\_grasp} or a receptive region for \texttt{pre\_inside}; Fig.~\ref{fig:reward_overview}(a)) and a joint-space distance for \texttt{seat}. For predicates not naturally represented by such a distance, we use a predicate-specific physical measure. Specifically, we evaluate \texttt{grasp} using CHORD's force-closure reward in 6D wrench space~\citep{chord}, augmented with a grasp-quality measure~\citep{lynch2017modern} to ensure grasp stability (see Appendix~\ref{app:rewards} for details and discussion).

With a reward defined for each predicate, we construct the task reward from the stage-wise scene graph. As the task progresses to a new stage, the reward for the newly introduced predicate is added to the task reward (Fig.~\ref{fig:reward_overview}(b)).
Because each predicate has a fixed reward definition, we  apply the same construction across different tasks. Fig.~\ref{fig:reward_overview}(c) shows this shared construction for direct and tool-use manipulation tasks: although they involve different entities and predicate sequences, their task rewards are assembled from the same predicate-level reward functions.

\textbf{Initial-state distribution from the stage-wise scene graphs.}
While predicate rewards guide progress from a state, the stage-wise scene graph defines the initial-state distribution by determining which parts of the task the policy can practice during training. 
Since each stage graph $\mathcal G_k$ fixes relations but not poses or grasps, we construct $\rho_0$ by using it as a generative constraint. 
For geometric predicates, the same feature distance that provides their dense progress signal also serves as a placement rule: we position the source node relative to its target node so that the condition in Table~\ref{tab:predicates} holds, applying these rules along the edges of $\mathcal G_k$ (Fig.~\ref{fig:pipeline}, Appendix~\ref{app:reset}).
For the final stage, we first sample a state that satisfies the goal predicate and then perturb it into a near-goal state, since an already successful reset provides less useful learning signal.
Episodes start uniformly from such resets across all stages, so the policy practices late interactions without first reaching them from the task start, which reduces the exploration burden of long-horizon tasks~\citep{omnireset}.

For \texttt{grasp}, the placement rule is grasp synthesis. Instead of fixing a single grasp, we place the hand at one of many multi-finger grasps generated by Dexonomy~\citep{dexonomy} within the object's grasp region. The policy thus encounters diverse hand--object configurations during training(Appendix~\ref{app:grasp_synth}). 
Each grasp also includes the finger command $q_{\mathrm{cmd}}$ that closes the hand on the object, which we restore at initialization so the grasp holds from the first step.
Before a sampled state enters $\rho_0$, we instantiate it in simulation and keep it only if it passes our checks for scene collision and reset stability, using grasps certified beforehand under external forces (Appendix~\ref{app:reset_verify}).

\subsection{Zero-shot Sim-to-Real Transfer}
\label{sec:sim2real}
We train the policy with PPO~\citep{ppo} using an asymmetric actor--critic~\citep{pinto2017asymmetric}, in which the critic receives privileged observations and the actor receives only the scene-graph-grounded observation available on the real robot. To enable zero-shot transfer, we randomize physical properties such as object mass and friction during training (Appendix~\ref{app:training_details}) and address the sim-to-real gap in perception and control. 
On the perception side, we perturb the actor observations during training with object-pose noise, perceived-size noise, and observation delay. At deployment, we estimate object poses from two RGB-D cameras using FoundationPose++~\citep{FoundationPose-plus-plus} and fuse the estimates with depth-consistency weights that down-weight unreliable views (Appendix~\ref{app:pose_fusion}).
On the control side, the policy outputs desired palm and fingertip velocities, which a QP low-level controller~\citep{lee2026learning} maps to joint velocities while enforcing kinematic and deployment-time safety constraints.
This lets us adjust joint-velocity limits and add collision-avoidance constraints at deployment without retraining the policy. Appendix~\ref{app:QP_controller} provides further details on the QP controller and demonstrates a real-world hardware failure prevented by the QP constraints.

\section{Experiments}
\label{sec:experiment_setup}

In this section, we evaluate \algoname in simulation and on real hardware. Our experiments address three questions: 
\begin{itemize}[leftmargin=2.5em, itemsep=0pt, topsep=1pt]
    \item[\textbf{(Q1)}] Does \algoname generalize to initial and goal configurations not shown in the single human video?
    \item[\textbf{(Q2)}] Does \algoname apply across diverse dexterous embodiments with different kinematics?
    \item[\textbf{(Q3)}] Which design choices of \algoname are critical to its performance?
\end{itemize}

\subsection{Experimental Setup} 
\noindent\textbf{Tasks.}
We evaluate \algoname on five manipulation tasks that span direct object manipulation and tool use (Fig.~\ref{fig:real_world_results}). The direct-manipulation tasks (Doll, Can, Stamp) require the robot to grasp an object and move it to a target configuration, while the tool-use tasks (Hammer, Sweep) require the robot to grasp and manipulate a tool to interact with another object. 

\noindent\textbf{Hardware Setup.}
We conduct all real-world experiments on a 6-DoF UR3 robot arm equipped with a 20-DoF Wuji 1 dexterous hand. Two Intel RealSense D435i RGB-D cameras capture the workspace from complementary viewpoints. We estimate object poses from these observations using FoundationPose++~\citep{FoundationPose-plus-plus}.

\noindent\textbf{Baselines.}
We compare our \algoname against two prior methods that learn dexterous manipulation from a single human video. These methods differ in what information they extract from the human demonstration and how they use it for policy learning.
{Do as I Do}~\citep{doasido} reconstructs the 4D hand--object interaction and retargets the recovered human hand motion to the robot through trajectory optimization. 
CHORD~\citep{chord} uses the demonstrated object-state evolution as an imitation reward and augments it with contact-wrench guidance that encourages the robot to reproduce the demonstrated trajectories.

\subsection{Does \algoname Generalize Beyond the Human Video?}
\label{exp:generalization}

\begin{figure}[t]
  \centering
  \includegraphics[width=\linewidth]{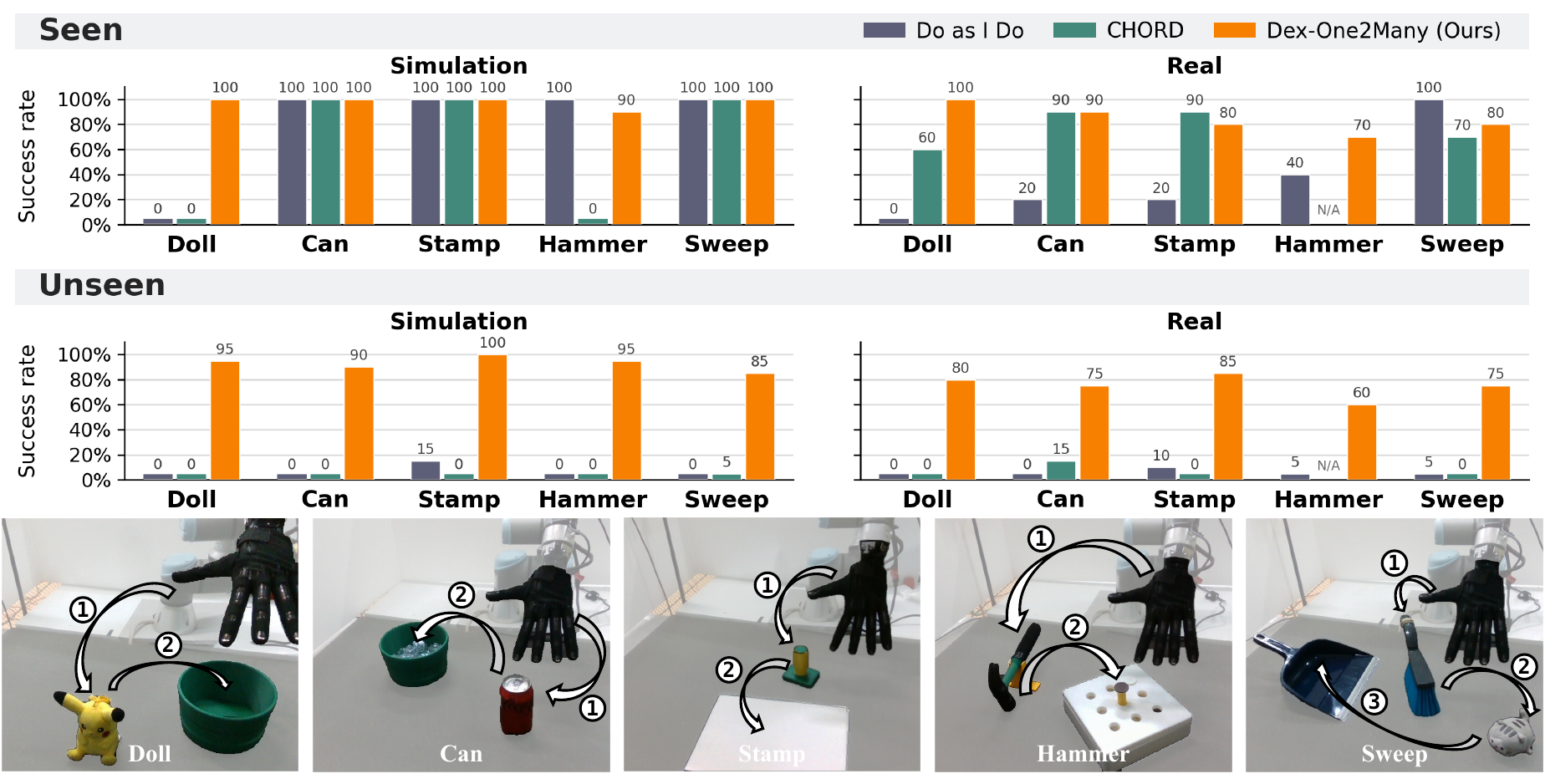}
  \caption{\textbf{Generalization beyond the human video in simulation and the real world.} Success rates are reported for configurations shown in the input human video (\emph{Seen}, top) and configurations with different initial and goal poses not shown in the video (\emph{Unseen}, bottom), in both simulation and the real world. The bottom row illustrates the five evaluated tasks; N/A denotes an experiment unavailable due to a hardware failure.
  }
  \label{fig:real_world_results}
  \vspace{-5pt}
\end{figure}

\noindent\textbf{Evaluation protocol.}
We evaluate generalization in both simulation and the real world using the same \emph{Seen}/\emph{Unseen} protocol. For each method and task, we evaluate 30 episodes in simulation and 30 trials on the real robot: 10 in \emph{Seen} configurations and 20 in \emph{Unseen} configurations.
The \emph{Seen} configurations reproduce the initial and goal poses shown in the input human video, whereas the \emph{Unseen} configurations use different initial and goal poses not present in the video. All evaluation episodes start with an empty hand and require the policy to execute the full task from reaching to task completion. We deploy real-world policies zero-shot, without real-world data or fine-tuning.

\noindent\textbf{Results.}
Figure~\ref{fig:real_world_results} compares performance on \emph{Seen} and \emph{Unseen} configurations in simulation and the real world. On \emph{Seen} configurations, the baselines can achieve high success on several tasks, showing that execution-level guidance extracted from the human video can reproduce the demonstrated task configuration. 
The difference becomes pronounced on \emph{Unseen} configurations. Across the evaluated baselines, success remains at or below 15\% in both simulation and the real world, whereas \algoname achieves 85--100\% success in simulation and 60--85\% on the real robot. 

\begin{figure}[!t]
  \centering
  \includegraphics[width=\linewidth]{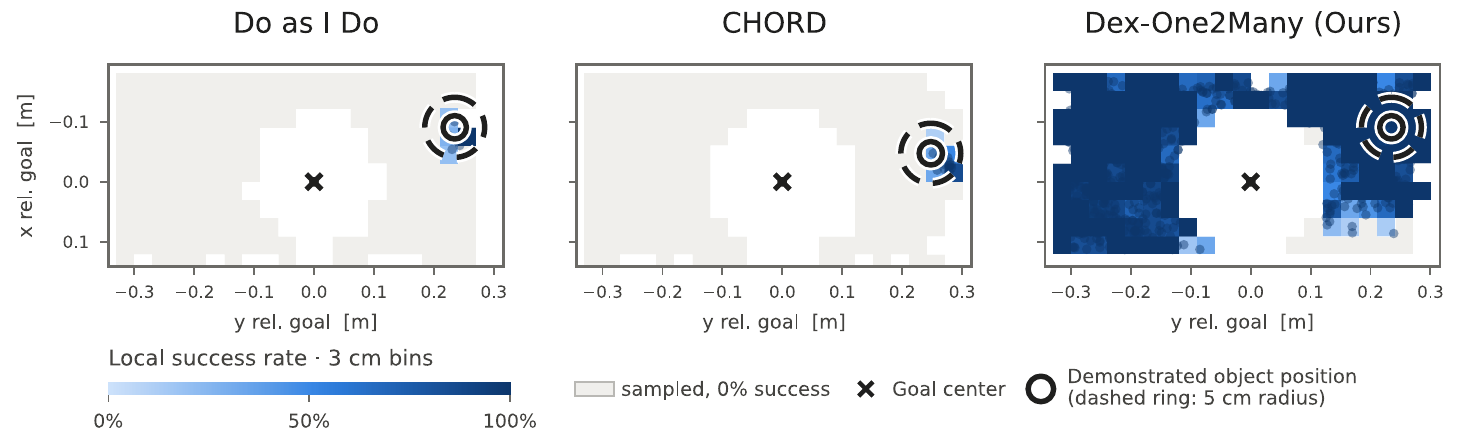}
  \caption{\textbf{Generalization to unseen initial object positions on the Can task.} In simulation with the goal fixed, \algoname succeeds across a broader range of initial positions than the baselines.
  }
  \label{fig:diverse_initial_can_landscape}
  \vspace{-7pt}
\end{figure}

Figure~\ref{fig:diverse_initial_can_landscape} further isolates generalization to unseen initial object positions on the Can task. With the goal fixed, the baselines succeed primarily near the demonstrated object position, whereas \algoname maintains high success across a substantially broader range of the workspace. This shows that the performance gap is not limited to a particular set of evaluation configurations, but reflects broader spatial generalization beyond the demonstrated execution. Additional results across the remaining tasks are provided in Appendix~\ref{app:initial_generalization}.

The baselines derive execution-level guidance from the demonstrated hand or object motion, which is tied to the execution observed in the video. In contrast, \algoname uses stage-wise scene graphs as generative constraints for diverse task-consistent training states, allowing the policy to learn the same task structure across different configurations beyond the demonstrated execution.

\begin{figure}[t]
    \centering
    \includegraphics[width=\linewidth]{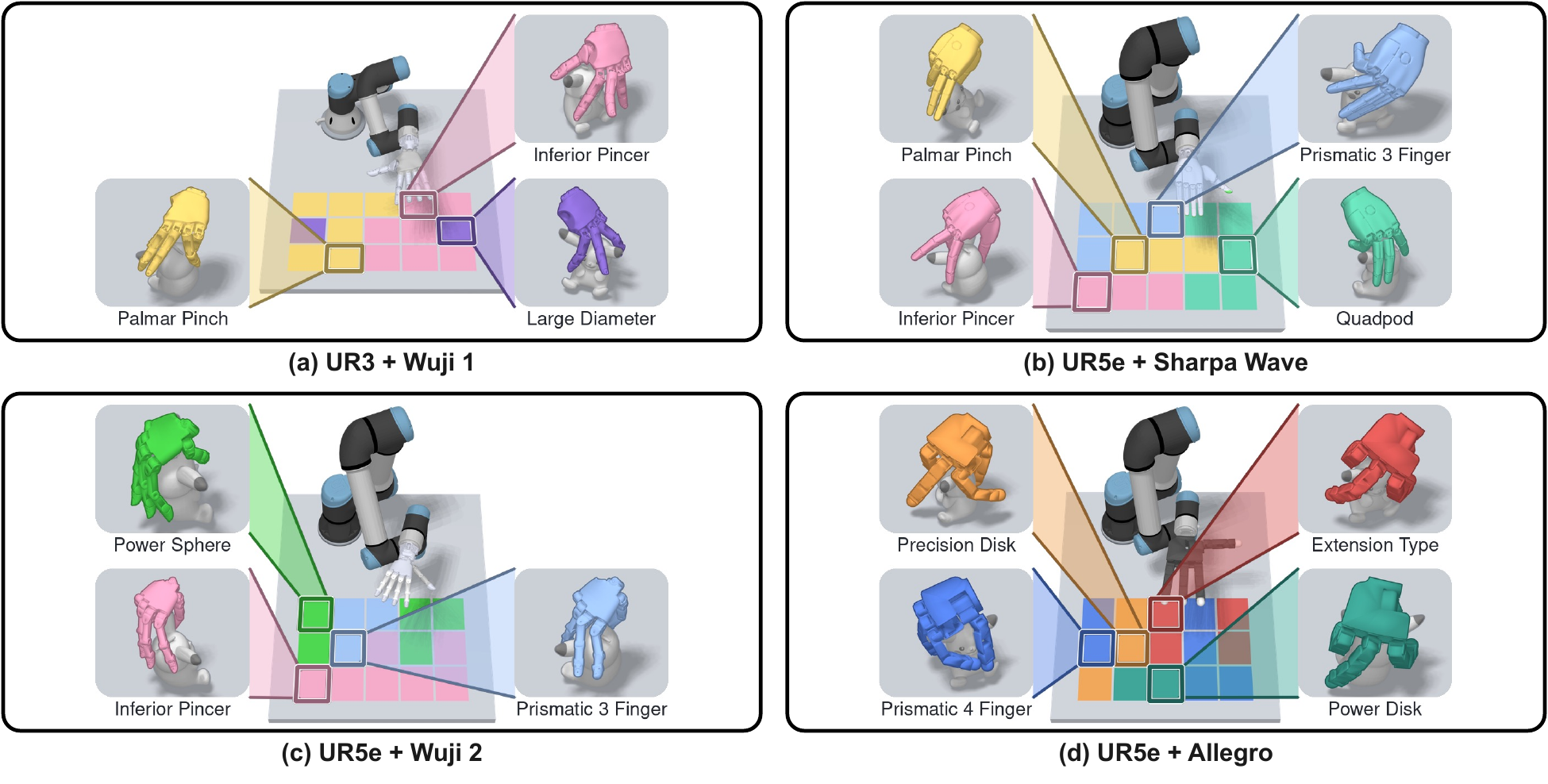}
    \caption{
    \textbf{Diverse grasp strategies across dexterous embodiments and object configurations on the Doll task.} Across the 33-type grasp taxonomy~\citep{feix2015grasp}, different embodiments learn different grasp types, and the same embodiment adapts its grasp across object configurations.
    }
    \label{fig:diverse_grasp_sim}
    \vspace{-12pt}
\end{figure}

\subsection{Does \algoname Apply Across Diverse Dexterous Embodiments?}
\label{exp:cross-emb}

\noindent\textbf{Evaluation protocol.} 
We evaluate \algoname in simulation across four arm--hand embodiments with different kinematics: UR3 + Wuji 1, UR5e + Sharpa Wave, UR5e + Wuji 2, and UR5e + Allegro. For each embodiment, we only replace the robot URDF and synthesize a corresponding grasp set, keeping the task specification and all other components unchanged. We train a separate policy for each embodiment and report success rates over 500 evaluation episodes per task, with initial object positions sampled within the reachable part of $(x, y) \in [0.3, 0.7] \times [-0.17, 0.4]$\,m.

\begin{wraptable}{r}{0.34\columnwidth}
\vspace{-12pt}
\centering

\caption{
\textbf{Simulation performance across diverse dexterous embodiments.} Success rates (\%) over 500 episodes per task.
}
\label{tab:cross-embodiment}

\footnotesize
\setlength{\tabcolsep}{1.6pt}
\renewcommand{\arraystretch}{1.05}

\begin{tabular}{lcccc}
\toprule
&
\shortstack{UR3\\Wuji 1} &
\shortstack{UR5e\\Sharpa} &
\shortstack{UR5e\\Wuji 2} &
\shortstack{UR5e\\Allegro} \\
\midrule
Doll   & 88.4 & 98.4 & 97.2 & 97.2 \\
Can    & 93.6 & 97.4 & 95.2 & 92.2 \\
Stamp  & 98.6 & 96.6 & 97.0 & 98.8 \\
Hammer & 94.6 & 94.2 & 97.6 & 91.8 \\
Sweep  & 98.0 & 96.6 & 97.0 & 100.0  \\
\midrule
Mean   & 94.6 & 96.6 & 96.8 & 96.0 \\
\bottomrule
\end{tabular}

\vspace{-5pt}
\end{wraptable}

\noindent\textbf{Results.}
Table~\ref{tab:cross-embodiment} reports task-wise success rates and the mean across five tasks for four distinct dexterous embodiments. Despite substantial differences in arm--hand kinematics, \algoname achieves consistently high performance across all four tested embodiments, with mean success rates ranging from 94.6\% to 96.8\%. Appendix~\ref{app:additional_cross_embodiment} provides rollouts across all five tasks and multiple configurations. Figure~\ref{fig:diverse_grasp_sim} further shows that the learned execution is embodiment-specific. Under the same task specification, different hands adopt different grasp types, and the same hand can change its grasp as the object configuration changes. 
Rather than retargeting a fixed human grasp, \algoname exposes RL to diverse grasps synthesized for each morphology, and the policy learns which grasp suits each scene configuration.
Appendix~\ref {app:additional_diverse_grasps} provides additional grasp examples across embodiments and configurations.

\subsection{Which Design Choices Are Critical to \algoname?}
\label{exp:ablation}

\begin{figure}[!t]
    \centering
    \begin{minipage}[t]{0.45\linewidth}
        \centering
        \includegraphics[width=\linewidth]{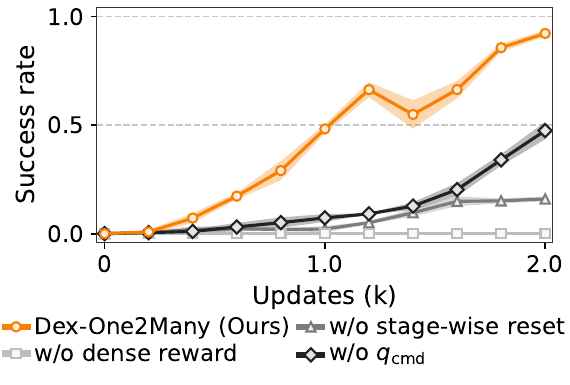}\\[-2pt]
        {\small (a) Component ablations}
    \end{minipage}\hfill
    \begin{minipage}[t]{0.45\linewidth}
        \centering
        \includegraphics[width=\linewidth]{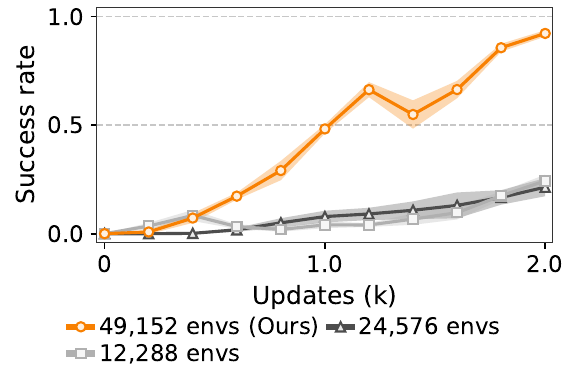}\\[-2pt]
        {\small (b) Training scale}
    \end{minipage}
    \caption{\textbf{Ablation studies.} (a) Removing each component. (b) Varying the number of environments.}
    \label{fig:ablation}
    \vspace{-10pt}
\end{figure}

\noindent\textbf{Evaluation protocol.}
We conduct all ablation studies on the Sweep task in simulation. Success is measured on full-task episodes from empty-hand starts. We study two aspects of \algoname: the contribution of its learning components and the effect of training scale. For component ablations, we remove one component at a time while keeping all other training settings unchanged. For the training-scale analysis, we vary only the number of parallel environments used for PPO training.

\noindent\textbf{Component ablations.}
From the human video, \algoname extracts a stage-wise scene graph that specifies the task relations at each stage.
We examine three design choices in how this task structure is used for policy learning.
\textbf{(1)} 
Removing stage-wise resets (w/o stage-wise reset) drops success after 2k PPO updates from over 90\% to about 16\%, showing that starting from intermediate stages is key to exploring multi-stage tasks.
\textbf{(2)} 
Removing predicate-derived rewards (w/o dense reward) keeps the success rate at 0\% throughout training, showing that dense feedback toward the required task relations is critical for guiding exploration.
\textbf{(3)} Finally, omitting the verified closing command (w/o $q_{cmd}$) lowers success to about 48\% after 2K PPO updates and induces frequent grasp slips and drops, showing that restoring active contact forces is vital for stable transitions across stages.

\noindent\textbf{Effect of training scale.}
Increasing the number of parallel environments substantially improves learning speed and final performance. With $49{,}152$ environments, \algoname reaches over 90\% success, whereas training with $24{,}576$ or $12{,}288$ environments remains below 25\% within the same number of PPO updates. These results show that large-scale parallel experience collection substantially benefits policy learning under our diverse reset distribution.

\section{Conclusion}
We presented \algoname, a real-to-sim-to-real framework for learning generalizable dexterous manipulation from a single human video. Rather than treating the demonstrated motion as an execution reference,  \algoname abstracts the video into a stage-wise scene graph that preserves the task structure while leaving object poses, goal poses, grasps, and motion paths free to vary. 
Scene-Graph-Grounded RL then uses this structure to construct diverse stage-wise reset states and predicate-derived rewards, making exploration tractable without inheriting the demonstrated trajectory.
Across five manipulation tasks, the resulting policies generalize to configurations not shown in the video and transfer zero-shot to a real multi-fingered hand.
These results suggest that extracting \emph{what} must be achieved from a human video, while allowing RL to discover \emph{how} to achieve it, provides a promising route from a single demonstration to diverse dexterous behaviors.

\noindent\textbf{Limitations and future work.}
Although \algoname substantially improves generalization across unseen configurations and applies to diverse hand morphologies, several limitations remain.
First, a sim-to-real gap remains because real-world deployment relies on estimated 6D object poses. Pose-estimation noise and differences between simulated and real object geometry can degrade policy performance, particularly during contact-rich interactions.
Second, our real-to-sim pipeline reconstructs only rigid objects, so articulated objects would require manually built assets.
Since a single video may not reveal the underlying articulation structure, such as joint locations, axes, and joint limits, automatic real-to-sim reconstruction remains difficult. Inferring these structural kinematics directly from unconstrained video is an important avenue for future work.

\section{Acknowledgment}

This work was partly supported by
the Institute of Information \& Communications Technology Planning \& Evaluation (IITP) grant funded by the Korea government (MSIT) [No.~RS-2021-II211343, Artificial Intelligence Graduate School Program (Seoul National University)],
the Technology Innovation Program (RS-2025-25453780, Development of a National Humanoid AI Robot Foundation Model for Multi-Task Applications) funded by the Ministry of Trade, Industry \& Energy (MOTIE, Korea),
and Samsung Electronics Robotics eXperience (RX).
We thank Jinu Pahk and Kangin Lee from Tommoro Robotics for providing the Wuji 1 hand used in our real-world experiments.

\clearpage
\bibliography{main}

\clearpage
\appendix

\addtocontents{toc}{\protect\setcounter{tocdepth}{3}}
\setcounter{tocdepth}{3}

\renewcommand{\contentsname}{Appendix}
\tableofcontents

\clearpage

\section{Additional Related Works}
\label{app: related_work}

\textbf{Demonstration-seeded RL in simulation.}
A parallel line makes multi-stage exploration tractable by injecting demonstration-derived information into RL in simulation.
Prior methods use this information as an action prior~\citep{rialto,resip}, as object-pose progress targets~\citep{xsim}, or to generate and refine task-specific reset and reward programs~\citep{video2policy}.
In each case, the demonstration guides exploration by specifying how the policy should act, progress, or be rewarded.
Instead, we use the demonstration to identify how object relations change at the predicate level encoded in the stage-wise scene graph over time.
These changes define the task stages, from which we construct diverse, physically valid reset states across all intermediate stages, including synthesized multi-finger grasps.
These stage-wise reset states allow episodes to start directly from intermediate stages of the task, reducing exploration burden. Furthermore, exposing the policy to multiple configurations results in a more generalizable policy. Thus, the video specifies the stage-wise relations the policy must achieve and their temporal order, rather than providing demonstrated actions, motion targets, or a reward program to follow.

\textbf{Task-Aware Grasp Synthesis and Learned Grasp Acquisition.}
TaskDexGrasp conditions static dexterous hand-pose synthesis on task semantics, while Dexterous Functional Grasping combines semantic affordances with a learned grasp-acquisition controller
~\citep{chen2024taskdexgrasp,agarwal2023dexfunc}. A broader line learns
generalizable grasp proposal, acquisition, or maintenance from synthetic
grasps and demonstrations ~\citep{xu2023unidexgrasp,wan2023unidexgrasp++,
yuan2026demograsp,grit,zhang2025robustdexgrasp}. These methods primarily target robust grasp acquisition across diverse objects, whereas our goal is to solve the full manipulation task. We use grasp synthesis to generate diverse reset states and train the policy via reinforcement learning to acquire the grasp and execute subsequent manipulation stages.

\textbf{Sim-to-Real Dexterous Manipulation.}
Simulation-based reinforcement learning has been successfully applied to real-world dexterous in-hand orientation~\citep{andrychowicz2020learning}. Subsequent work has extended this setting toward faster control~\citep{handa2023dextreme}, more accessible hardware, and tactile feedback~\citep{yang2025anyrotate}. 
Recent systems move toward longer-horizon interaction. SimToolReal~\citep{simtoolreal} trains a policy to move tools toward target poses and executes unseen tasks by tracking the full 6D tool trajectory extracted from a human video. Because this target pose sequence is assumed available during execution, and our task setting does not provide such trajectory information at execution time, the two settings are not directly comparable. Play2Perfect~\citep{play2perfect} and ADEPT~\citep{adept} instead focus on reusable dexterous pretraining: Play2Perfect learns from task-agnostic play before fine-tuning on downstream assembly tasks, while ADEPT pretrains generic object-reposing skills and adapts them to insertion and placement. These approaches complement to ours, as they study how pretraining supports downstream manipulation, whereas we focus on learning a new task from a single human video.

\textbf{Force Closure and Grasp-Quality Objectives.}
Force closure is widely used to evaluate and synthesize stable multi-finger grasps. Recent methods formulate differentiable force-closure objectives to generate physically stable dexterous grasps across objects and hand embodiments ~\citep{liu2021synthesizing,wang2023dexgraspnet,graspqp}.

Force-closure measures have also been used as reinforcement learning rewards. CHORD~\citep{chord} introduces a force-closure reward for learning dexterous manipulation when demonstration-derived contact guidance is unreliable. We adopt this force-closure term by augmenting it with a weakest-direction grasp-quality measure to encourage stable grasps. While concurrent work similarly uses grasp quality as a dense reward to strengthen weak wrench directions during in-hand manipulation~\citep{chen2026robust}, we augment the force-closure reward with a weakest-direction grasp-quality measure to acquire stable grasps as part of a multi-stage manipulation task.

\section{Task Abstraction as a Stage-wise Scene Graph}
\label{app:scene_graph}

This section details how we obtain the stage-wise scene graph from a human video. Sec.~\ref{app:predicate} lists the predicate vocabulary. Sec.~\ref{app:VLM} describes how a single VLM query converts the video into the stage-wise scene graph, including the input prompt, an example output, and the query time and cost. Sec.~\ref{app:asset_prepare} describes how we reconstruct object assets from the demonstration, estimate their physical properties, and localize their functional regions. The pretrained models used at each step are summarized in Table~\ref{tab:models_list}.

\begin{table}[!htbp]
\centering\footnotesize
\setlength{\tabcolsep}{4pt}
\caption{Pretrained models used in task abstraction, asset preparation, and real-world perception.}
\label{tab:models_list}
\begin{tabular}{@{}p{0.24\linewidth}p{0.22\linewidth}p{0.32\linewidth}l@{}}
\toprule
Step & Model & Input $\rightarrow$ Output & Sec. \\
\midrule
Scene-graph extraction & Gemini~3.1 Pro & video $\rightarrow$ entities, parts, stage sequence & \ref{app:VLM} \\
Instance segmentation & SAM~3.1 & first frame, \texttt{visual\_phrase} $\rightarrow$ object mask & \ref{app:asset_prepare} \\
Mesh reconstruction & SAM~3D Objects & RGB, mask, metric point map $\rightarrow$ metric mesh & \ref{app:asset_prepare} \\
Object pose tracking & FoundationPose++ & RGB-D stream, mesh $\rightarrow$ 6D pose & \ref{app:real} \\
Physical parameters & GPT-5 & category, size $\rightarrow$ mass, friction & \ref{app:functional_region} \\
Part localization & SAM~3 & rendered views, part name $\rightarrow$ part region & \ref{app:functional_region} \\
\bottomrule
\end{tabular}
\end{table}

\newcommand{\predtablesize}{\fontsize{8.0pt}{8.2pt}\selectfont}
\newcommand{\prednotesize}{\fontsize{7pt}{7.3pt}\selectfont}

\begin{table}[H]
\centering
\begin{threeparttable}

\caption{
\textbf{Predicate vocabulary.}
Geometric features, distance metrics, and constraints used to instantiate
the predicates in our tasks.
Paired pre-relation and terminal predicates share the same distance metric
and differ only in their constraints, except for
\texttt{pre\_grasp}/\texttt{grasp}.
}
\label{tab:predicates}

\predtablesize
\setlength{\tabcolsep}{2.5pt}
\renewcommand{\arraystretch}{1.15}

\begin{tabular*}{\linewidth}{
@{\extracolsep{\fill}}
l l l l l
@{}}
\toprule
\textbf{Predicate}
& \textbf{Reference $A$}
& \textbf{Target $B$}
& \textbf{Distance metric}
& \textbf{Constraint} \\
\midrule

\texttt{pre\_on\_top}
& \multirow{2}{*}{top support disk $\mathcal D_A$}
& \multirow{2}{*}{bottom point $p_B$}
& \multirow{2}{*}{$d_{\mathrm{top}}=d(p_B,\mathcal D_A)$}
& $\epsilon_{\mathrm{top}}<d_{\mathrm{top}}\le\varphi_{\mathrm{top}}$ \\

\texttt{on\_top}
&
&
&
& $d_{\mathrm{top}}\le\epsilon_{\mathrm{top}}$ \\

\addlinespace[2pt]
\cmidrule(lr){1-5}
\addlinespace[2pt]

\texttt{pre\_inside}
& \multirow{2}{*}{interior box $\mathcal B_A$}
& \multirow{2}{*}{root point $p_B$}
& \multirow{2}{*}{$d_{\mathrm{in}}=d(p_B,\mathcal B_A)$}
& $\epsilon_{\mathrm{in}}<d_{\mathrm{in}}\le\varphi_{\mathrm{in}}$ \\

\texttt{inside}
&
&
&
& $d_{\mathrm{in}}\le\epsilon_{\mathrm{in}}$ \\

\addlinespace[2pt]
\cmidrule(lr){1-5}
\addlinespace[2pt]

\texttt{pre\_contact}
& \multirow{2}{*}{center $p_A$}
& \multirow{2}{*}{working point $p_B^{\mathrm{work}}$}
& \multirow{2}{*}{$d_{\mathrm{con}}=\|p_A-p_B^{\mathrm{work}}\|_2$}
& $\epsilon_{\mathrm{con}}<d_{\mathrm{con}}\le\varphi_{\mathrm{con}}$ \\

\texttt{contact}
&
&
&
& $d_{\mathrm{con}}\le\epsilon_{\mathrm{con}}$ \\

\addlinespace[2pt]
\cmidrule(lr){1-5}
\addlinespace[2pt]

\texttt{pre\_seat}
& \multirow{2}{*}{goal $q_{\mathrm{goal}}$}
& \multirow{2}{*}{joint position $q_B$}
& \multirow{2}{*}{$d_{\mathrm{seat}}=|q_B-q_{\mathrm{goal}}|$}
& $\epsilon_{\mathrm{seat}}<d_{\mathrm{seat}}\le\varphi_{\mathrm{seat}}$ \\

\texttt{seat}
&
&
&
& $d_{\mathrm{seat}}\le\epsilon_{\mathrm{seat}}$ \\

\addlinespace[2pt]
\cmidrule(lr){1-5}
\addlinespace[2pt]

\texttt{pre\_grasp}
& grasp-region $p_A^{\mathrm{grasp}}$
& fingertips $P_B^{\mathrm{tip}}$
& $d_{\mathrm{pre\text{-}grasp}}
   =d_{\mathrm{tip}}(P_B^{\mathrm{tip}},p_A^{\mathrm{grasp}})$
& $\ell_g\le d_{\mathrm{pre\text{-}grasp}}\le u_g$ \\

\texttt{grasp}
& grasp set $\mathcal G_A$
& hand configuration $h_B$
& --
& $h_B\in\mathcal G_A$ \\

\bottomrule
\end{tabular*}

\begin{tablenotes}[flushleft]
\prednotesize
\item[]
\textit{Notation.}
$d(p,S)=\min_{s\in S}\|p-s\|_2$ denotes the Euclidean distance from a
point $p$ to a set $S$.
$\mathcal D_A$ and $\mathcal B_A$ denote the top support disk and interior
box of reference entity $A$, respectively.
For relation $c$, $\epsilon_c$ denotes its satisfaction tolerance, while
$\varphi_c$ denotes the outer sampling bound of the corresponding pre-relation.
$P_B^{\mathrm{tip}}$ denotes the fingertip positions, and
$d_{\mathrm{tip}}$ denotes the fingertip-to-grasp-region-center distance
used for \texttt{pre\_grasp}.
$\ell_g$ and $u_g$ denote its lower and upper sampling bounds.
$\mathcal G_A$ denotes the verified grasp set for object $A$, and
$h_B$ denotes the hand configuration.
\end{tablenotes}

\end{threeparttable}
\end{table}

\subsection{Predicate Vocabulary}
\label{app:predicate}
Table~\ref{tab:predicates} lists the predicates in our vocabulary, which cover common relations in tabletop manipulation. Each geometric predicate also has an approach form, denoted by the prefix \texttt{pre-}, which holds while the relation is being approached but not yet established.

\subsection{VLM Query}
\label{app:VLM}

\begin{figure}[!tbh]
  \centering
  \includegraphics[width=\linewidth]{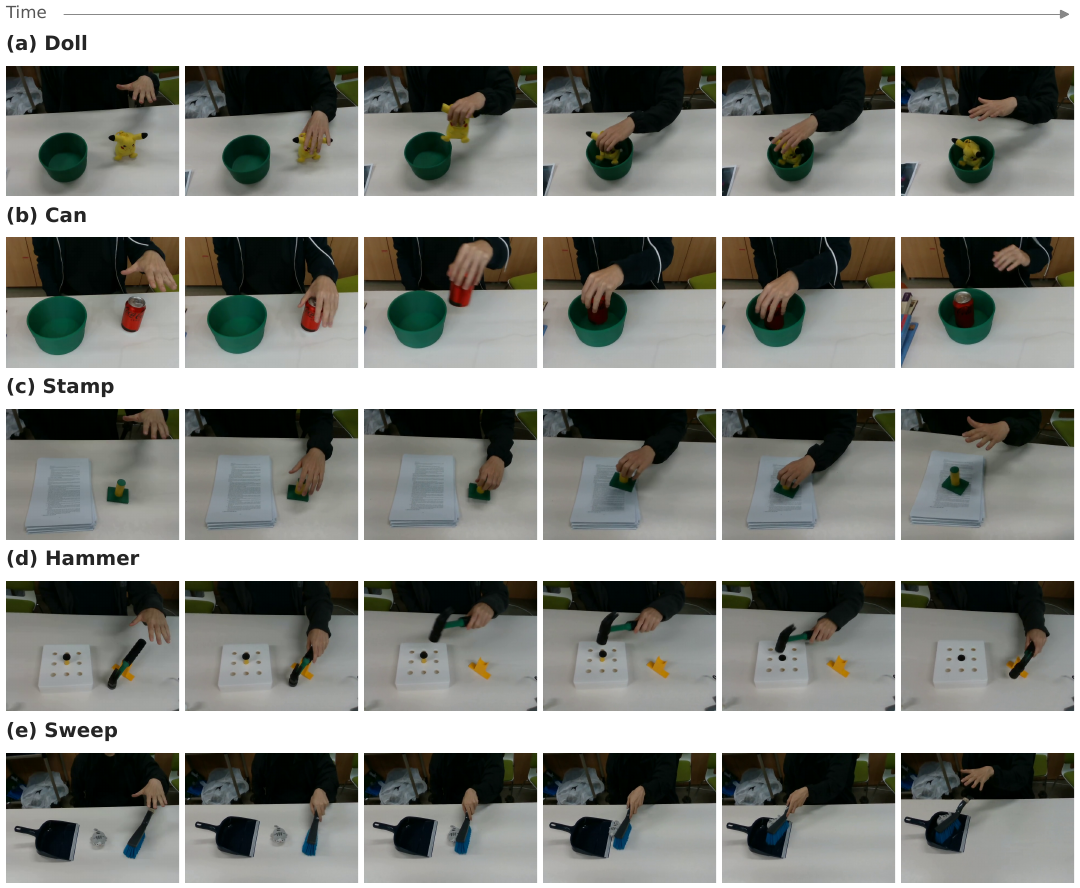}
  \caption{\textbf{Single human demonstrations for the five tasks.} Each row shows the human video used to specify one task, with frames ordered from left to right over time.}
  \label{fig:human_video_sequence}
\end{figure}

Given the single human demonstration of each task (Fig.~\ref{fig:human_video_sequence}), recorded at $640\times480$, we query Gemini~3.1 Pro (\texttt{gemini-3.1-pro-preview})~\citep{google_gemini_3_1_pro_2026} with the entire video at medium media resolution; videos under 20,MB are sent inline, and larger ones through the file-upload API. The prompt (Listing~\ref{lst:app_prompt}) is identical for all tasks: the only task-specific input is the video, and the vocabulary in the prompt is the one in Table~\ref{tab:predicates}. A single query takes 12--21s and costs less than \$0.05, so extracting all five task graphs takes under 90s and \$0.20 in total (Table~\ref{tab:app_vlm_cost}).

\begingroup 
\fontsize{8}{9.5}\selectfont
\setlength{\tabcolsep}{4pt}
\setlength\LTleft{0pt} 
\setlength\LTright{0pt} 

\begin{longtable}{@{\extracolsep{\fill}}lp{0.28\linewidth}p{0.16\linewidth}cp{0.29\linewidth}l@{}}
\caption{\textbf{Task suite.} Object roles: \emph{held} is grasped by the robot, \emph{site} receives another object, \emph{acted-on} is manipulated only through a tool, and \emph{fixture} holds an object at the start. The predicate chain extracted by the VLM defines $K$ stages, each with its own reset distribution. Objects marked with $^\dagger$ use a CAD mesh instead of a reconstructed one (Appendix~\ref{app:asset_prepare}).}
\label{tab:task_suite} \\ 

\toprule
Task & Description & Objects (role) & $K$ & \shortstack[l]{Predicate chain\\(stage $1 \rightarrow K$)} & Type \\
\midrule
\endfirsthead 

\caption*{\textbf{Task suite.} (Continued)} \\
\toprule
Task & Description & Objects (role) & $K$ & \shortstack[l]{Predicate chain\\(stage $1 \rightarrow K$)} & Type \\
\midrule
\endhead 

\bottomrule
\endfoot
\endlastfoot

Doll   & \raggedright Pick up a Pikachu plush toy and place it inside an orange pot. & \raggedright pikachu (held),\newline pot$^\dagger$ (site)
       & 5 & \begin{tabular}[t]{@{}l@{\;}l@{}}
             & $\varnothing$ \\
             $\rightarrow$ & \texttt{pre-grasp}(hand,\,pikachu) \\
             $\rightarrow$ & \texttt{grasp}(hand,\,pikachu) \\
             $\rightarrow$ & \texttt{pre-inside}(pikachu,\,pot) \\
             $\rightarrow$ & \texttt{success:}\textbf{\texttt{inside}}(pikachu,\,pot)
             \end{tabular}
       & direct \\
\addlinespace[5pt]
Can    & \raggedright Pick up the coke can and place it in a bucket. & \raggedright can (held),\newline bucket$^\dagger$ (site)
       & 5 & \begin{tabular}[t]{@{}l@{\;}l@{}}
             & $\varnothing$ \\
             $\rightarrow$ & \texttt{pre-grasp}(hand,\,can) \\
             $\rightarrow$ & \texttt{grasp}(hand,\,can) \\
             $\rightarrow$ & \texttt{pre-inside}(can,\,bucket) \\
             $\rightarrow$ & \texttt{success:}\textbf{\texttt{inside}}(can,\,bucket)
             \end{tabular}
       & direct \\
\addlinespace[5pt]
Stamp  & \raggedright Grasp the stamp and press it down onto the stack of papers. & \raggedright stamp$^\dagger$ (held),\newline paper (site)
       & 5 & \begin{tabular}[t]{@{}l@{\;}l@{}}
             & $\varnothing$ \\
             $\rightarrow$ & \texttt{pre-grasp}(hand,\,stamp) \\
             $\rightarrow$ & \texttt{grasp}(hand,\,stamp) \\
             $\rightarrow$ & \texttt{pre-on\_top}(stamp,\,paper) \\
             $\rightarrow$ & \texttt{success:}\textbf{\texttt{on\_top}}(stamp,\,paper)
             \end{tabular}
       & direct \\
\addlinespace[5pt]
Hammer & \raggedright Use a hammer to strike a peg inserted into a block. & \raggedright hammer$^\dagger$ (held),\newline peg$^\dagger$ (acted-on),\newline block$^\dagger$ (site), \newline holder$^\dagger$ (fixture)
       & 6 & \begin{tabular}[t]{@{}l@{\;}l@{}}
             & \texttt{fixed\_to}(hammer,\, holder) \\
             $\rightarrow$ & \texttt{pre-grasp}(hand,\,hammer) \\
             $\rightarrow$ & \texttt{grasp}(hand,\,hammer) \\
             $\rightarrow$ & \texttt{pre-contact}(hammer,\,peg) \\
             $\rightarrow$ & \texttt{contact}(hammer,\,peg) \\
             $\rightarrow$ & \texttt{success:}\textbf{\texttt{seated}}(peg,\,block)
             \end{tabular}
       & tool-use \\
\addlinespace[5pt]
Sweep  & \raggedright Sweep the toy cat into the dustpan using the brush. & \raggedright brush (held),\newline toy (acted-on),\newline dustpan (site)
       & 7 & \begin{tabular}[t]{@{}l@{\;}l@{}}
             & $\varnothing$ \\
             $\rightarrow$ & \texttt{pre-grasp}(hand,\,brush) \\
             $\rightarrow$ & \texttt{grasp}(hand,\,brush) \\
             $\rightarrow$ & \texttt{pre-contact}(brush,\,toy) \\
             $\rightarrow$ & \texttt{contact}(brush,\,toy) \\
             $\rightarrow$ & \texttt{pre-inside}(toy,\,dustpan) \\
             $\rightarrow$ & \texttt{success:}\textbf{{inside}}(toy,\,dustpan)
             \end{tabular}
       & tool-use \\
\end{longtable}
\endgroup

\begin{table}[!htbp]
\centering
\caption{Structure of the VLM output and the use of each field in our pipeline. Indented entries are fields of the object above them, and \texttt{[]} marks a list with one entry per object, mount, or stage.}
\label{tab:vlm_fields}
\scriptsize
\setlength{\tabcolsep}{4pt}
\begin{tabularx}{\linewidth}{@{}l >{\raggedright\arraybackslash}X >{\raggedright\arraybackslash}X@{}}
\toprule
\textbf{Field} & \textbf{Content} & \textbf{Used for} \\
\midrule
\multicolumn{3}{@{}l}{\texttt{task}} \\
\quad \texttt{name}, \texttt{description} & Task name and one-line summary & Generated first to condition the subsequent fields; not used by the pipeline \\
\midrule
\multicolumn{3}{@{}l}{\texttt{objects[]}} \\
\quad \texttt{id} & Identifier of each task-relevant object & Nodes; key for all per-object data \\
\quad \texttt{category} & Object class & Hint in the physical-parameter query \\
\quad \texttt{visual\_phrase} & Text description of the object's appearance & Object segmentation \\
\quad \texttt{roles} & Role of the object (e.g., held, site) & Scene slots in reset generation and RL \\
\quad \texttt{upright} & Whether the object is normally kept upright & Restricts sampled stable poses to the upright class \\
\quad \texttt{grasping\_part}, \texttt{working\_part} & Names of functional parts & Text prompts for functional-region localization \\
\quad \texttt{working\_direction} & How a tool meets its target (down or horizontal) & Contact placement \\
\midrule
\multicolumn{3}{@{}l}{\texttt{scene\_graph}} \\
\quad \texttt{initial\_state[]} & Relations that already hold in the first frame & First stage graph; kept in the resets of later stages until the related object moves \\
\quad \texttt{stages[]} & One predicate per stage, in temporal order & Stage sequence for backward reset generation; the last stage sets the near-goal target \\
\quad \texttt{goal\_condition} & Predicates that must all hold at success & RL success test (first predicate) \\
\bottomrule
\end{tabularx}
\end{table}

\newpage
The VLM returns a JSON object following a fixed response schema, whose fields and their use in our pipeline are summarized in Table~\ref{tab:vlm_fields}. Listing~\ref{lst:app_sweep_json} shows the output for the Sweep task, in which a brush is used to push a toy cat into a dustpan, and Table~\ref{tab:task_suite} summarizes the resulting objects, roles, and stage sequences for all five tasks.  Table~\ref{tab:task_suite} lists the objects, their roles, and the stage sequence extracted by the VLM from each video.

\begin{promptbox}[label=lst:app_prompt]{VLM prompt for stage-wise scene graph extraction. \texttt{\{predicate\_list\}} is filled with the predicates of Table~\ref{tab:predicates}.}
You are analyzing a human demonstration video for robot learning (real2sim).
The human hand in the video corresponds to a ROBOT END-EFFECTOR. Refer to it as "ee".
Do NOT list the hand as an object; it is the agent, not a manipulable object.

Extract, grounded ONLY in what is visually observable:

1. TASK: concise name and one-line description.

2. OBJECTS: the manipulable objects of interest (NOT the hand). For each: a short id,
   a category, a visual_phrase, and roles.
   category vs visual_phrase — these are DIFFERENT and both required:
   - "category": the clean object class used to search an asset library ("kettle", "mug",
     "knife"). A superordinate is fine here.
   - "visual_phrase": what you would type into an open-vocabulary SEGMENTATION model to
     select exactly this object in the video frame. 2-4 words, concrete and appearance-
     bearing: colour and/or material plus the specific noun ("red enamel kettle", "white
     ceramic mug", "wooden cutting board"). NEVER an abstract superordinate ("container",
     "object", "item", "tool") — those ground to nothing. If two objects in the frame share
     a noun, the phrase must be what tells them apart.
   Do NOT include the workspace surface itself (worktable, desk, floor, walls) — it
   is part of the robot cell, not an object. But DO include a fixed object that holds
   another object off that surface (a knife holder, a dish rack, a tray): it is not
   manipulated, yet the scene is wrong without it.
   Likewise include every object that another object is, or ends up, "seated" in
   (inserted, driven, or screwed into it: the bottle a cork is pressed into, the
   socket a plug is pushed into), even if the hand never touches it.
   roles — assign ALL that apply, judged from what visibly happens in the video:
   - "held":     the hand grasps and holds this object at some point
                 (both directly-placed objects and tools count).
   - "site":     another object comes to rest ON or IN this object
                 (plate under food, cutting board, container receiving something).
   - "acted_on": a HELD object acts on this object through contact
                 (food being cut, dough being rolled, a window being wiped).
   - "fixture":  this object is already holding another object up when the demo
                 begins, and the hand never touches it (a holder, a rack, a tray).
   "site" and "fixture" differ in WHEN: a site receives an object as a result
   of the manipulation, a fixture is holding one before it starts. The plate a
   cake is placed on is a site; the holder the knife rests in is a fixture.
   A pot that is carried and later receives food is ["held", "site"].
   Also determine per object:
   - "upright": true if the object is normally stored/used standing upright
     (cup, bottle, vase, jar → true; plate, cutting board, spatula, knife → false).
   - "grasping_part": for "held" objects, the part the hand grasps.
   - "working_direction": for "held" tool-use objects, how the working part moves
     against the target at the moment of contact — "down" if the tool is raised and
     brought down onto it (pestle, potato masher, cookie cutter), "horizontal" if the tool is pushed
     sideways across a surface (squeegee, spatula, rolling pin). This decides the
     contact pose the sampler builds; getting it wrong produces a tool that slides sideways
     when it should come down. null if not tool-use.
   - "working_part": for "held" objects that physically contact another
     object to do work (tool-use), the functional surface that does the work.
     null if the held object is placed directly (not tool-use).

   PART NAMING — these two strings are fed verbatim to an open-vocabulary
   segmenter as text prompts. Name the SMALLEST region that actually carries
   the part's role, and describe ONLY that region.

   Write "<attribute> <part> of the <object>" with EXACTLY ONE attribute, and
   that attribute must belong to the region you are naming — not to anything
   attached to it. If the handle is the part, describe the handle alone: its
   own colour, its own material. Say nothing about the head, the shaft, the
   body, or the object's overall look, even when those features are the most
   obvious thing in the demo. Every word describing a neighbouring region is
   an invitation for the mask to spread into that region.

   Keep the <object> slot bare — the plain object noun, never "the red mug"
   or "the wooden-handled knife".

   NEVER enumerate. If the region itself shows two colours, pick the one that
   covers most of it — do not write both. A colour the region shares with a
   neighbouring region is fine: name the region as it actually looks. Never
   borrow a colour from an adjacent part just to stand apart from it.
       good: "blue handle of the screwdriver"  (grip region only)
             "black handle of the spatula"     (same black as the blade — still
                                                right, that is what it looks like)
             "rubber blade of the squeegee"    (blade only, not the frame)
             "metal blade of the knife"
             "black grip of the pan"
       bad:  "handle"                          (no attribute — bleeds outward)
             "silver handle of the ladle"      (silver is the neck; the grip
                                                the hand holds is black — a colour
                                                borrowed from the neighbour)
             "black handle of the silver pan"  (attribute on the object —
                                                mask jumps to the body)
             "wooden handle of the metal knife"  (describes the blade too)
             "blue and red handle of the screwdriver" (enumeration — matches
                                                       the whole screwdriver)
             "blue rubber ergonomic handle of the screwdriver"  (piled attributes)
   Use only attributes actually visible in the demo; never invent a colour.
   If no attribute applies to the region on its own, write just
   "<part> of the <object>". null if not applicable.

3. SCENE_GRAPH: express it with PREDICATES ONLY (do not write free-form sentences),
   in three parts: initial_state, goal_condition, stages.

   initial_state: what is already mounted on what in the FIRST frame, before the
   hand moves. Most demos start with everything simply lying on the table — in
   that case return an empty list. Emit an entry only when an object begins the
   demo **held off the table by another object** in a definite resting pose: a
   ladle hanging on its rack, a part seated in a tray, a bit sitting in a holder.
   The only predicate here is "fixed_to": a is the object being held, b is the
   thing holding it. Nothing else — this list is not about what the hand does.
   - arg 'a' must be an object that LATER MOVES (it appears in the
     stages). An object that never moves needs no entry: it is simply part
     of the scene.
   - arg 'b' is the holder. Give it the "fixture" role in OBJECTS.
   - Do not describe an object resting directly on the table or the ground.
   - Do not describe a part that is already inserted, driven, or screwed into
     another object. That is a joint, not a mount, and it is not expressed here.

   goal_condition and stages — allowed predicates: {predicate_list}.
   - goal_condition: the set of predicates that must ALL hold at success (goal state).
   - stages: the relations that become established during the demo, one per entry,
     in the temporal order in which they are established. List only relations that
     become TRUE; do NOT add entries for approaching, reaching, or moving toward
     something — approach phases are derived automatically from these relations.
     Use "ee" as arg 'a' for grasp.
   Be precise about when each relation is actually established (e.g. distinguish
   real contact from the tool merely hovering near the target).
   AGENCY TRANSFER: when a source acts on a target through "grasp" or "contact", the
   target becomes arg 'a' of the subsequent stages — it is the thing that moves
   next, not the original actor. (ee grasps spatula → spatula becomes arg 'a'; spatula contacts
   pancake → pancake becomes arg 'a' for the move onto the plate.)
   Choose predicates from the list; never invent new ones.
   - "inside" vs "seat": use "seat" ONLY when the object ends up DRIVEN/PRESSED INTO
     the other one so the two become a single rigid assembly (a cork pressed into a
     bottle, a plug pushed into a socket, a drawer pushed shut). Use "inside" when the
     object merely ends up within an open container and could be lifted straight out
     (fruit placed in a basket, an item dropped in a bin).
   (grasp->grasp, contact->contact, place->on_top/inside/seat).

Do not hallucinate objects not present. Return only valid JSON matching the schema.

\end{promptbox}

\begin{jsonbox}[label=lst:app_sweep_json]{VLM output for the Sweep task}
{
  "task": {
    "name": "sweep toy into dustpan",
    "description": "Use a brush to sweep a small plush toy into a dustpan."
  },
  "objects": [
    {
      "id": "brush",
      "category": "brush",
      "visual_phrase": "blue brush",
      "roles": ["held"],
      "upright": false,
      "grasping_part": "black handle of the brush",
      "working_part": "blue bristles of the brush",
      "working_direction": "horizontal"
    },
    {
      "id": "dustpan",
      "category": "dustpan",
      "visual_phrase": "blue dustpan",
      "roles": ["site"],
      "upright": false,
      "grasping_part": null,
      "working_part": null,
      "working_direction": null
    },
    {
      "id": "toy",
      "category": "plush toy",
      "visual_phrase": "grey cat plush",
      "roles": ["acted_on"],
      "upright": true,
      "grasping_part": null,
      "working_part": null,
      "working_direction": null
    }
  ],
  "scene_graph": {
  "initial_state": [],
  "stages": [
    {"step": 1, "predicate": "grasp",   "args": {"a": "ee",    "b": "brush"}},
    {"step": 2, "predicate": "contact", "args": {"a": "brush", "b": "toy"}},
    {"step": 3, "predicate": "inside",  "args": {"a": "toy",   "b": "dustpan"}}
  ],
  "goal_condition": [
    {"predicate": "inside", "args": {"a": "toy", "b": "dustpan"}}
  ]
}
}
\end{jsonbox}

\begin{table}[ht]
\centering\scriptsize
\caption{VLM query cost per task.}
\label{tab:app_vlm_cost}
\begin{tabular}{lrrr}
\toprule
Task & Video [s] & Latency [s] & Cost [\$] \\
\midrule
Doll   & 7.07 & 21.2 & 0.045 \\
Can    & 5.30 & 11.5 & 0.024 \\
Stamp  & 4.97 & 12.0 & 0.026 \\
Hammer & 4.00 & 18.7 & 0.039 \\
Sweep  & 5.77 & 18.5 & 0.037 \\
\midrule
Total  & 27.11 & 81.9 & 0.171 \\
\bottomrule
\end{tabular}
\end{table}

\subsection{Asset Preparation and Functional Regions}
\label{app:asset_prepare}

We reconstruct each task-relevant object from the first frame of the demonstration; the full video is used only by the VLM query. The object's visual phrase from the VLM output serves as the text prompt for SAM 3.1 \citep{carion2026sam}, which returns the object's instance mask. Given the RGB image, the mask, and the metric point map from the depth image of the same frame, SAM 3D Objects \citep{chen2026sam} reconstructs the object mesh. For objects marked in Table~\ref{tab:task_suite}, we instead use an available CAD mesh.

Each mesh, reconstructed or CAD, is post-processed into a watertight simulation mesh and a set of convex pieces (Table~\ref{tab:mesh_post}). The convex pieces serve as collision geometry for grasp synthesis, and the watertight mesh serves as the collision mesh in Isaac Sim~\citep{mittal2025isaaclab}.
\begin{table}[H]
\centering\scriptsize
\setlength{\tabcolsep}{3pt}
\caption{Mesh post-processing.}
\label{tab:mesh_post}
\begin{tabular}{@{}p{0.2\linewidth}p{0.22\linewidth}p{0.54\linewidth}@{}}
\toprule
Step & Tool & Parameters \\
\midrule
Convex decomposition & CoACD~\citep{wei2022coacd} & concavity threshold 0.05; resolution 2000; MCTS 20 nodes, 150 iterations, depth 3; $rv_k=0.3$; manifold preprocessing only for non-manifold input (resolution 50); hull merging on; no hull-count limit; minimum part volume $10^{-10}$ (\texttt{-mv}); random seed from the system clock \\
Watertight remesh & CoACD manifold mode (OpenVDB~\citep{museth2013vdb}) & applied to the normalised mesh; dual-marching-cubes level set 0.1 \\
Simplification & ACVD~\citep{valette2004acvd} & 2000 vertices, gradation 1.5, manifold output \\
\bottomrule
\end{tabular}
\end{table}

\subsection{Physical Properties and Functional Region Localization}
\label{app:functional_region}

The VLM query provides only semantic attributes of each object (Table~\ref{tab:vlm_fields}); physical parameters and functional regions are obtained as follows.

\paragraph{Physical parameters.}
Mass, mass range, and friction coefficient are estimated after scene-graph extraction by a separate text-only language-model query (\texttt{gpt-5}~\citep{singh2026openaigpt5card}), which receives the id, category, and measured longest axis of each object(Listing~\ref{lst:app_phys_prompt}).

Here, \texttt{\{obj\_lines\}} holds one line per object: its id, category, and measured longest axis in cm. During RL, the object mass is sampled from $[m-\Delta m,\, m+\Delta m]$ and the static friction from $[0.9\mu,\, 1.1\mu]$ with a floor of $0.05$, and the dynamic friction is set to $0.9$ times the static friction (Table~\ref{tab:dr}). Physical verification (in Sec.~\ref{app:reset_verify}) use the worst case of these ranges, namely the maximum mass and the minimum friction.
\begin{promptbox}[label=lst:app_phys_prompt]{Prompt for physical-parameter estimation}
This is a domain-randomization setup for a robot simulation. The objects
below were 3D-scanned from real instances, so the longest-axis size is already fixed by
measurement -- do NOT re-estimate the size.

For each object, answering as "an object of that category whose measured size is that value",
give:
1. mass_g: typical mass at that size (g, empty / dry weight).
2. mass_range_g: +/- variation of the mass (g).
3. friction_coeff: static friction coefficient against a flat surface
   (plastic ~0.3, wood ~0.4, rubber ~0.8, ceramic ~0.4, metal ~0.3, fabric ~0.5).
Objects:
{obj_lines}

Answer in JSON, keeping each id exactly as given.
\end{promptbox}

\paragraph{Grasp and working regions.}
We localize each part name from the VLM output on the object mesh in three steps (Table~\ref{tab:app_regions}). We first render the mesh from 34 viewpoints around the object and segment the part in each view with SAM~3, using the part name as a text prompt. We then project the 2D masks back onto the mesh and keep the vertices that are detected in most of the views in which they are visible. Finally, the grasping region is taken as the largest connected component, and the working region is fitted with planar surfaces that represent the tool's working faces. To keep tool heads out of the grasping region, masks of generic tool-head concepts (e.g., blades and bristles) are removed from it. The grasping region restricts grasp synthesis and provides the approach target of the reward, and the working surfaces are used by the contact sampler and the tool reward.

\begin{table}[t]
\centering\scriptsize
\setlength{\tabcolsep}{3pt}
\caption{Localization of grasping and working regions.}
\label{tab:app_regions}
\begin{tabular}{@{}p{0.2\linewidth}p{0.76\linewidth}@{}}
\toprule
Step & Setting \\
\midrule
Views & 34 views: elevations $-35^\circ, 10^\circ, 35^\circ, 60^\circ$ $\times$ 8 azimuths ($45^\circ$ steps), plus top ($85^\circ$) and bottom ($-85^\circ$); rendered twice (textured and uniform grey); $768{\times}768$ px, vertical field of view $45^\circ$, camera distance $2.6\times$ the half bounding-box diagonal \\
Detection & SAM~3; the part string is one text concept; instance score floor 0.15, mask threshold 0.5 \\
Negative concepts & grasping only: \emph{blade, cutting edge, spike, bristles}; their masks are dilated by 12\,px and removed \\
Fusion & a vertex is visible in a view if its depth is within $2\%$ of the bounding-box diagonal of the z-buffer; score $=$ positive votes / visible views, over views with at least one detection; vertex kept if seen in $\ge 2$ views and score $>0.3$ \\
Transfer to mesh & each simulation-mesh face takes the maximum score of the nearest triangle of the reconstructed mesh; face kept if score $\ge 0.25$; grasping region: largest connected component \\
Outputs & per-face mask and object-frame bounding box (centre, extents) of the grasping region; for grasp synthesis on the table, the mask is restricted to faces whose lowest vertex is $\ge 5$\,mm above the table \\
Working surfaces & working-region faces clustered by adjacency with normal deviation $\le 5^\circ$; area-weighted plane fit per cluster; clusters with RMS residual $\le 6\times10^{-5}$\,m and area $\ge 90\%$ of the largest are kept (centre, normal, half-extents) \\
\bottomrule
\end{tabular}
\end{table}

\paragraph{Support patches.}
Support patches are the upward-facing flat regions of a mesh on which another object can rest. Faces whose normal has a vertical component above $0.99$ (within about $8^\circ$ of vertical) are grouped by height with a 5\,mm tolerance, and groups with less than $5\%$ of the largest area are dropped. Each patch stores its height, centre, and radius.

\newpage
\section{Scene-graph-grounded Reinforcement Learning}
\label{app:rl}

In this section, we make an detail explanation for Sec. \ref{sec:rl}. 

\subsection{Observations}
\label{app:observations}

\begin{wrapfigure}{r}{0.45\textwidth}
    \centering
    \includegraphics[width=\linewidth]{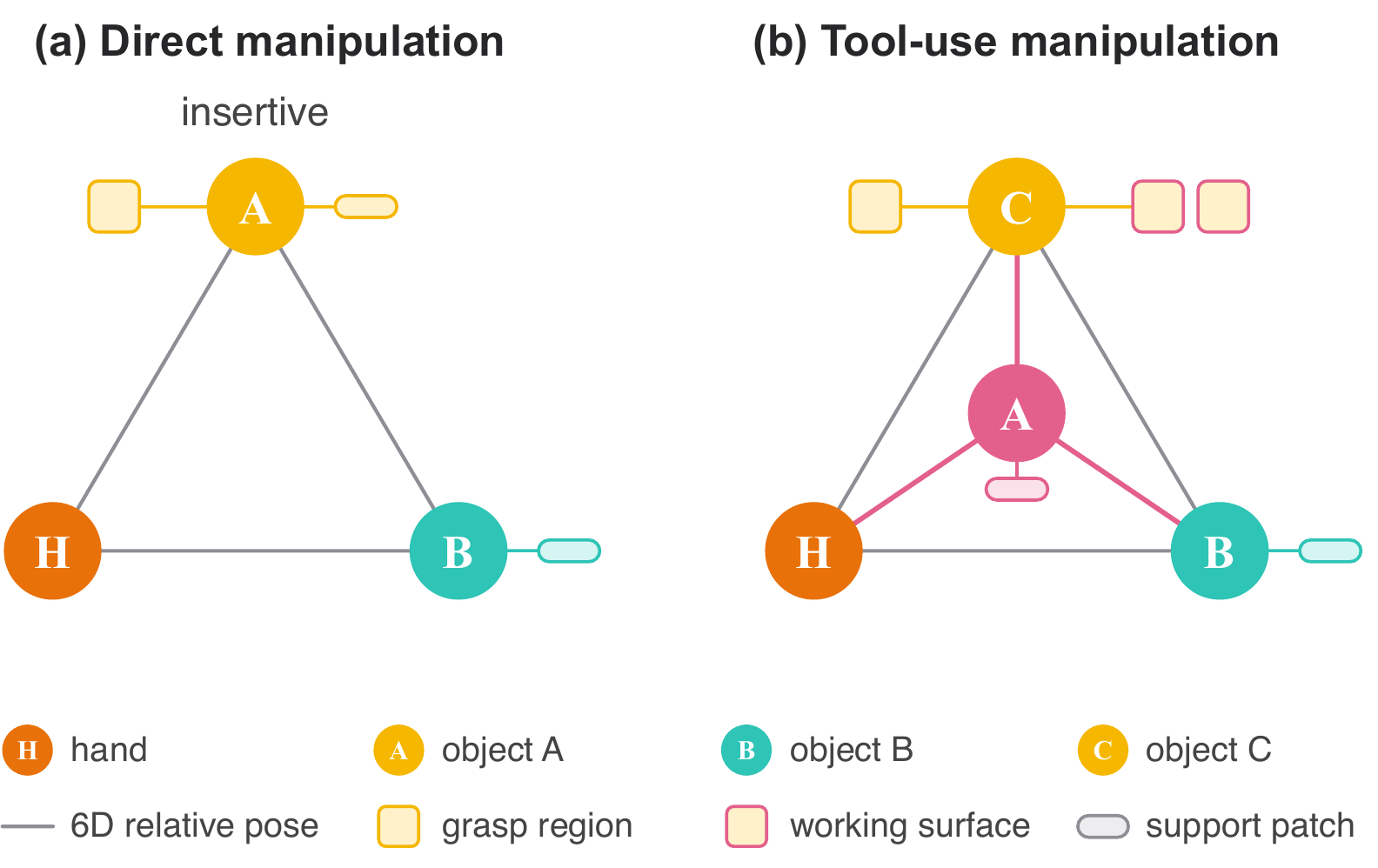}
    \caption{ 
    \textbf{Scene-graph-grounded observation.} 
    The policy observation is constructed from scene-graph nodes and their relative 6D-pose edges for (a) direct and (b) tool-use manipulation. 
    }
    \label{fig:observation}
    \vspace{-15pt}
\end{wrapfigure}

\paragraph{Scene-graph-grounded observation construction.}
The scene graph identifies the task-relevant entities, their roles, relations,
and functional regions. We map the demonstrated hand to the robot end-effector
and assign the remaining objects according to their roles in the task.
For direct manipulation, the observation contains the end-effector, the
manipulated object, and the receptive object. For tool-use manipulation, the
held object is instead the tool, and an additional acted-on (interactive) object is introduced (Fig.~\ref{fig:observation}).

Object and functional-part nodes provide their local geometry, such as object size, grasp regions, and working surfaces. Functional parts are associated with the hand or another object according to their role in the interaction. We represent a grasp region by its center and bounding-box extents, and a working surface by its center, unit normal vector of its surface, and the extent of the surface.

Each scene-graph edge contributes a relative 6D pose between its connected
entities. Importantly, this relative pose is defined according to the predicate encoded by the edge: the predicate determines which geometric features of the two entities are used, following the same feature definitions used for the predicate distance functions.
Thus, an edge may relate object centers, functional regions, or other
predicate-specific features rather than simply the two object centers.
All node, part, and edge features are concatenated to form the policy
observation.

\subsection{Rewards}
\label{app:rewards}
The \texttt{grasp} predicate is evaluated in 6D wrench space rather than by a
Cartesian distance. Following CHORD~\citep{chord}, we approximate the friction
cone at each contact with a set of edge forces, each inducing a 6D primitive
wrench consisting of force and torque. We then evaluate how well these
primitive wrenches support a set of sampled directions in wrench space.
Let $\sigma_{j,t}$ denote the support along sampled direction $j$ at time $t$.
CHORD's force-closure reward measures the fraction of directions whose support
exceeds a threshold:
\begin{equation*}
    r_{\mathrm{fc}}(s_t, a_t)
    =
    \frac{1}{B}
    \sum_{j=1}^{B}
    \mathbb{I}[\sigma_j > \epsilon].
\end{equation*}
where $B$ is the number of sampled wrench directions and $\epsilon$ is the
minimum support threshold.

In our setting, we use $r_{\mathrm{fc}}$ to guide grasp acquisition without
a retargeted hand trajectory or virtual-object controller~\citep{dexmachina}. Directional coverage alone, however, can increase for contacts that do not form a stable grasp; for example, the policy may push the brush against the toy cat instead of first grasping the brush. We therefore augment $r_{\mathrm{fc}}$ with the Ferrari--Canny grasp quality~\citep{ferrari1992planning}, which measures support along the weakest wrench
direction:
\begin{equation}
    Q_t = \min_j \sigma_{j\,t} .
    \label{eq:grasp_quality}
\end{equation}
The resulting grasp reward is
\begin{equation}
    r_{\mathrm{grasp}}(o_t, a_t)
    =
    r_{\mathrm{fc}}(o_t, a_t)
    +
    \mathbb{I}\!\left[N_{\mathrm{finger}, t}\ge2\right]
    \max(Q_t,0),
    \label{eq:r_force}
\end{equation}
where $N_{\mathrm{finger},t}$ is the number of distinct fingers in contact with the object. The additional quality term favors stable multi-finger grasps by encouraging support even along the weakest wrench direction.

\textbf{Progress rewards from spatial predicates.}
For predicates whose relations can be expressed as a Cartesian distance,
we define dense progress rewards from reductions in that distance. Following~\citet{simtoolreal}, we measure progress relative to the smallest distance reached so far in the episode. For $\texttt{pre-grasp}(A,B)$, let $d_{AB,t}$ denote the current distance and $d^{*}_{AB,t}$ the smallest distance reached up to time $t$. We define
\begin{equation*}
    r_{\mathrm{pre-grasp},t}(A,B)
    =
    \max \;\bigl(d^{*}_{AB,t-1} - d_{AB,t}, 0\bigr) ,
\end{equation*}
so that reward is produced only when the policy makes new progress toward the relation. Using improvement over the best distance reached so far avoids rewarding stationary behavior: once the distance stops decreasing, the progress reward becomes zero.
We use the same progress construction for the other distance-based
predicates. For \texttt{contact} predicate, we additionally include an alignment term
between the tool's working region and the desired direction of interaction.

\textbf{Task reward composition.}
We next describe how the reward terms are combined during training.
We use a grasp indicator $\mathbb{I}_{\mathrm{hold},t}$ to determine whether the object is currently lifted by a valid grasp. Using the same grasp-quality criterion as above, we define
\begin{equation}
    \mathbb{I}_{\mathrm{hold},t}
    =
    \mathbb{I}\!\left[N_{\mathrm{finger},t}\ge2\right]
    \mathbb{I}\!\left[Q_t>0\right]
    \mathbb{I}\!\left[z_t-z_0 > h_{\mathrm{lift}}\right],
    \qquad h_{\mathrm{lift}}=5\,\mathrm{cm},
    \label{eq:hold_indicator}
\end{equation}
Thus, $\mathbb{I}_{\mathrm{hold},t}$ is active only while at least two distinct fingers are in contact, the grasp provides positive support along its weakest wrench direction and the target is lifted above $h_\mathrm{lift}$.

Let $g$ denote the first stage whose scene graph contains a
\texttt{grasp} predicate, and let $r_{i,t}$ denote the reward driven by the predicates associated with stage $\mathcal G_i$.
At the grasp stage, we additionally use a lift reward before the hold condition is satisfied:
\begin{equation}
    r_{g,t}
    =
    r_{\mathrm{grasp},t}
    +
    \bigl(1-\mathbb{I}_{\mathrm{hold},t}\bigr)
    r_{\mathrm{lift},t}.
    \label{eq:r_grasp_stage}
\end{equation}
The complete task reward is then
\begin{equation}
    r_t
    =
    \sum_{i \le g} r_{i,t}
    +
    \mathbb{I}_{\mathrm{hold},t}
    \sum_{i > g} r_{i,t}
    +
    r_{\mathrm{success},t}
    +
    r_{\mathrm{reg},t},
    \label{eq:task_reward_full}
\end{equation}
where rewards through the grasp stage are always available, while rewards
associated with subsequent manipulation stages are activated only when the
object satisfies the hold condition in Eq.~\ref{eq:hold_indicator}.
Here, $r_{\mathrm{success},t}$ is the terminal success bonus, and
$r_{\mathrm{reg},t}$ penalizes large actions and joint velocities.

\subsection{Reset Generation}
\label{app:reset}
\begin{figure}[!t]
  \centering
  \includegraphics[
    width=\linewidth,
  ]{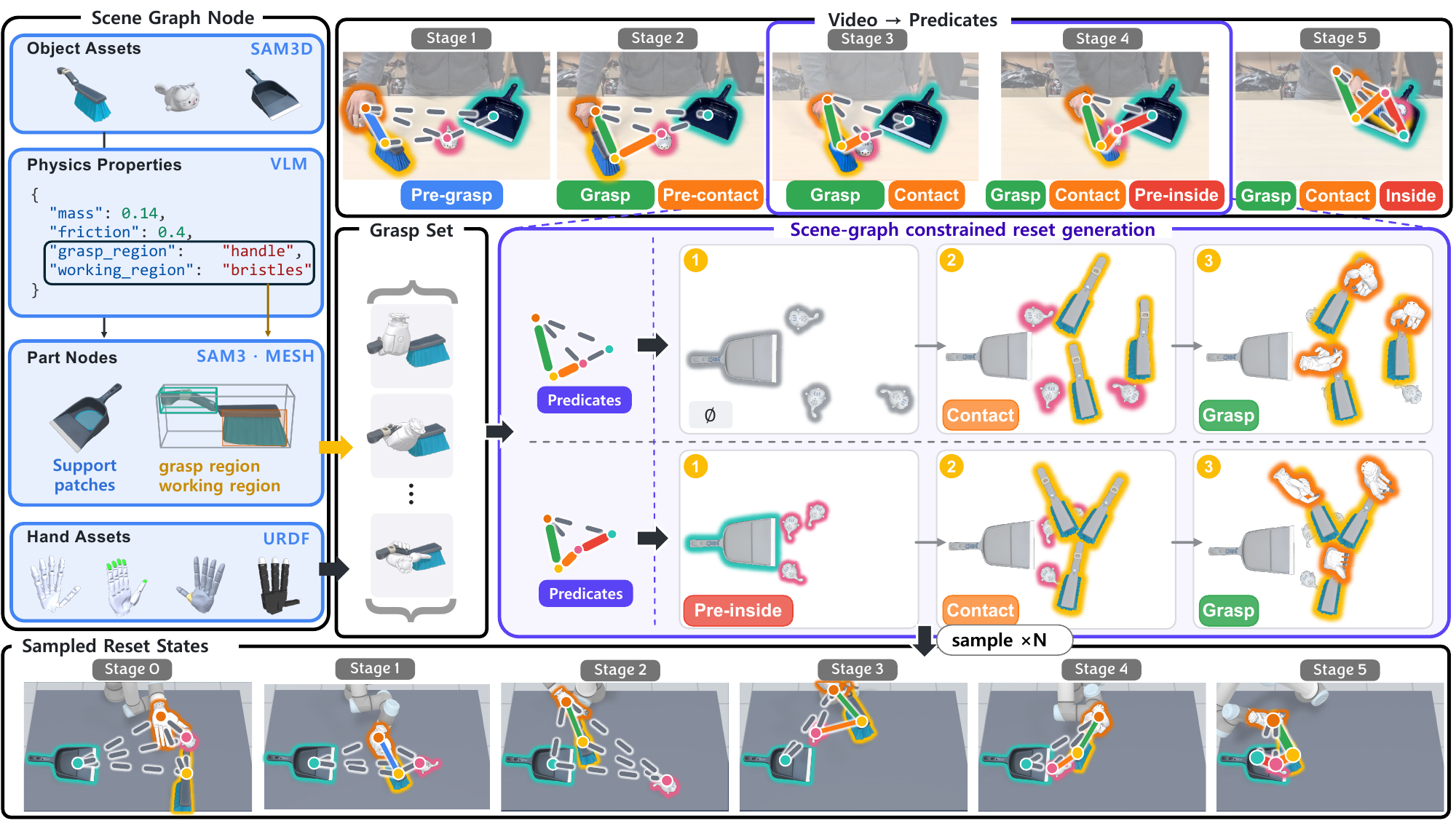}
  \caption{
  \textbf{Scene-graph-constrained reset generation.} Given the stage-wise scene graph and the assets, we instantiate each task stage as diverse reset states. The object and hand assets provide geometry and physical properties, while functional part nodes identify regions such as grasp and working regions. Together, these are used to construct a grasp set, restricted to the specified grasp region when available. For each stage, predicates in the scene-graph act as generative constraints: geometric predicates determine where related entities can be sampled, while the \texttt{grasp} predicate samples a valid hand--object configuration from the grasp set. Repeating this process produces diverse simulation reset states that satisfy the relations specified by each stage graph.
}
  \label{fig:pipeline}
\end{figure}

This section details how we construct the stage-wise initial-state distribution $\rho_0$ of Sec.~\ref{sec:rl}. Fig.~\ref{fig:pipeline} summarizes the process. 
Each stage graph $\mathcal{G}_k$ specifies which relations must hold at that point of the task but leaves object poses and grasps free, so many geometrically distinct states satisfy it. We generate such states in two steps. First, we sample states that satisfy the predicates of the stage graph, using each predicate as a rule for placing one node relative to another (Sec.~\ref{app:reset_sampler}); the \texttt{grasp} predicate is instantiated from a set of synthesized dexterous grasps (Sec.~\ref{app:grasp_synth}). Second, we keep only the states that remain usable for the rest of the task and are physically valid in simulation (Sec.~\ref{app:reset_verify}). We retain 10{,}000 verified states per stage, which form its reset distribution. The resulting states are therefore not random configurations of the scene, but diverse instantiations of the same task structure.

\subsubsection{Predicate Samplers and Placement Order}
\label{app:reset_sampler}
Each predicate in Table~\ref{tab:predicates} is defined on geometric features of its two nodes, and the same features determine where one node can be placed relative to the other. For example, \texttt{on\_top} restricts the bottom point of the source object to the support patch of the target, and \texttt{inside} restricts the root point of the source object to the interior box of the target. We use these conditions as placement rules: to satisfy a predicate, we sample the pose of its source node within the region that its target node and the predicate define.

To construct a full stage, we resolve its predicates in dependency order along the edges of $\mathcal{G}_k$. The anchor, a node that is the target but not the source of any predicate, is placed first. We then follow the edges from the anchor and place each source node relative to its already placed target. Finally, unconstrained nodes, which appear in no predicate of the stage, are sampled in the remaining free space.
Fig.~\ref{fig:pipeline} illustrates this process for two stages of the Sweep task. The stage in the top row contains \texttt{contact}(brush, toy cat) and \texttt{grasp}(hand, brush). The toy cat is the anchor, so it is placed first; the brush is then placed relative to it so that its bristles touch the toy cat, and the hand is placed on the brush handle at a grasp from $\mathcal{H}_{\mathrm{grasp}}$ (Sec.~\ref{app:grasp_synth}). The dustpan appears in no predicate of this stage and is sampled last in the remaining free space. The stage in the bottom row additionally contains \texttt{pre-inside}(toy cat, dustpan), so the dustpan becomes the anchor: it is placed first, the toy cat is placed relative to it without being inside it, and the brush and the hand follow as above.
 For the final stage, we first sample a state that satisfies the goal predicate and then perturb it into a near-goal state, following OmniReset~\citep{omnireset}, since an already successful reset provides less useful learning signal.

\subsubsection{Grasp Synthesis}
\label{app:grasp_synth}
A valid grasp is critical for stage-wise resets: every stage that contains \texttt{grasp} presupposes a securely held object, so an invalid grasp invalidates all of them. We synthesize grasp candidates with Dexonomy~\citep{dexonomy}, which takes a robot hand, an object mesh, and a grasp type from a 33-type taxonomy, and returns diverse grasps of that type. When the object has a grasp region (Appendix~\ref{app:functional_region}), we restrict the synthesis to that region, so that, for example, the brush is grasped at its handle rather than its bristles. For each grasp, we also store the finger command $q_{\mathrm{cmd}}$ that closes the hand on the object and restore it at initialization together with the hand configuration. The corresponding \texttt{pre-grasp} state is obtained by linearly interpolating the finger joints toward the open configuration, to 30--90\% of the grasp configuration, and retracting the hand by 1--5\,cm along its approach axis.

\FloatBarrier

\begin{figure}[!tbp]
  \centering
  \includegraphics[width=\linewidth]{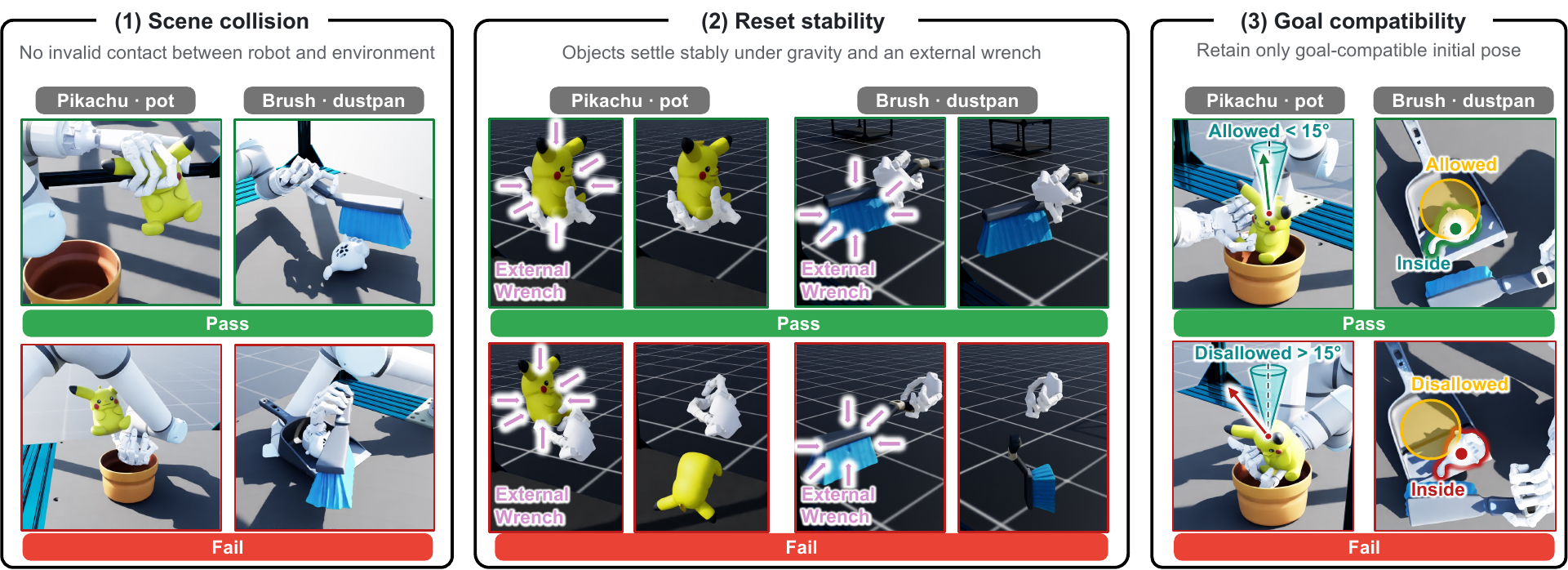}
  \caption{\textbf{Physical verification of reset candidates in the Sweep task.} (1) and (2) (top: pass, bottom: fail), and (3) is enforced over grasp types. (1) \emph{Scene collision}: the robot makes no invalid contact with the table or other objects; the failing candidate collides with the dustpan. (2) \emph{Reset stability}: objects settle under gravity and the grasped object stays in the hand under external wrenches; the failing grasp drops the brush. (3) \emph{Goal compatibility}: the goal relation remains attainable with the candidate grasp; the failing candidate cannot bring the toy cat inside the allowed region of the dustpan.}
  \label{fig:app_verification}
\end{figure}

\subsubsection{Transition Feasibility and Physical Validity}
\label{app:reset_verify}
Satisfying the predicates of one stage does not guarantee that a state remains usable later in the task. We therefore check compatibility at the grasp and stage levels, and verify every state physically.

\emph{Grasp level.} A grasp that is stable in isolation may still place the hand in a configuration from which the rest of the task cannot be completed. Some grasps are valid when the object is picked up but fail in later stages. For example, when an object must be placed \texttt{inside} a narrow container, a grasp that wraps the whole object holds it securely at pickup, but brings the fingers into collision with the container at the goal. Because every stage after the grasp inherits the same grasp, such a grasp makes all of these stages unusable. We therefore select grasps by their compatibility with the whole task rather than by stability alone. For each grasp type, we instantiate its candidates in every stage that contains \texttt{grasp} and apply the physical checks below. We rank the grasp types by the fraction of candidates that pass across all such stages, and the candidates of the top-$n$ types ($n=4$) form the grasp set $\mathcal{H}_{\mathrm{grasp}}$. Grasp types that hold the object but cannot attain the goal relation, such as the one in Fig.~\ref{fig:app_verification}~(3), receive a low pass rate and are excluded. The resulting grasp set is therefore conditioned on the entire task, not only on how well each grasp holds the object.

\emph{Stage level.} Sampling each stage independently can produce states that satisfy their own stage graph yet cannot lead to the next stage. For example, a \texttt{pre-grasp} state might place a pan with its handle pointing away from the robot, where no verified grasp of the next stage is reachable. Such states are valid locally but ill-posed for the task as a whole. We prevent them by generating the stages backward, starting from the goal stage. After the states of stage $k$ are verified, we carry the poses of their anchor objects and their grasps into stage $k-1$, and resample only the remaining entities under the predicates of stage $k-1$. Each state of stage $k-1$ is thus constructed around a physically verified state of stage $k$ rather than in isolation, so the policy can always progress from one stage to the next.

\emph{Physical validity.} Every candidate state is instantiated in simulation with the full robot and must pass two checks (Fig.~\ref{fig:app_verification}). (1) \emph{Scene collision}: the robot makes no invalid contact with the table or other objects. (2) \emph{Reset stability}: at the end of a 6\,s episode, all objects have been at rest for the last 2\,s, and in grasped states the object is still held by at least two hand links. Goal compatibility is enforced at the grasp level rather than per state: grasp types whose candidates cannot attain the goal relation fail these checks in the goal stage and are removed by the ranking above (Fig.~\ref{fig:app_verification}~(3)). All checks use the worst-case physical parameters of the domain randomization, the maximum mass and the minimum friction (Appendix~\ref{app:functional_region}), so that verified states remain valid across the randomized range. Grasp robustness is verified once per grasp before reset generation: with only the hand and the object, the grasp must hold the object under gravity in six directions and under an external force of $\min(0.5\,m_{\max}g,\,0.5\,\text{N})$ with a 0.02\,m lever along each of the six axes. The pass rate of each grasp type across all grasped stages provides the ranking at the grasp level above, and only states that pass all checks are retained and propagated backward to the previous stage.

\subsection{Hierarchical decomposition with a QP low level.}
\label{app:QP_controller}
Zero-shot sim-to-real deployment can expose the robot to hardware-specific constraints that are difficult to anticipate during simulation training. For example, excessive joint velocities may lead to unsafe motions, while contacts that are benign in simulation can damage the physical system. In our setup, contact between the Wuji 1 hand coupler and the table can disengage the coupler from the robot. 

To accommodate deployment-time safety constraints without retraining the policy, we separate learned task-space control from constrained joint-space execution. Following prior work~\citep{xie2022glide,
gangapurwala2022rloc,bang2024rl}, we let the RL policy act in task space and delegate kinematic resolution to a model-based low-level controller. Specifically, we adopt the QP velocity-IK formulation of~\citet{lee2026learning}, which maps the policy's palm and fingertip velocity commands to joint velocities subject to kinematic and safety constraints.

\begin{figure}[!t]
  \centering
  \includegraphics[width=\linewidth]{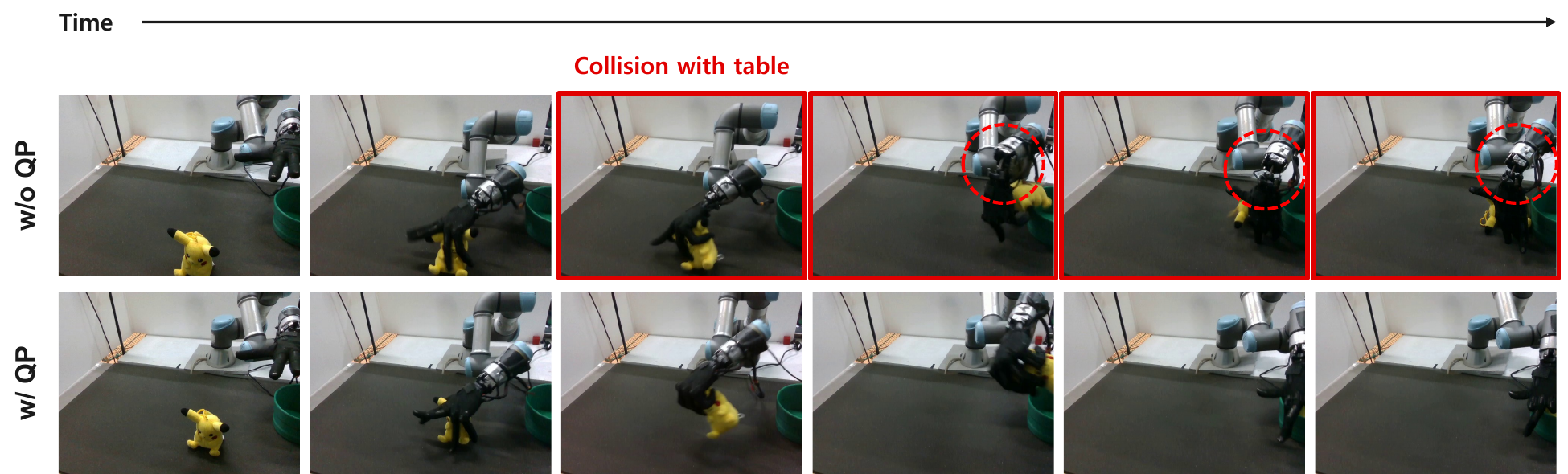}
  \caption{\textbf{QP constraints address a hardware-specific sim-to-real gap.}
Hand--table contact is benign in simulation because coupler disengagement is not modeled, but the same contact can disengage the Wuji~1 hand coupler on the physical robot.
Without the QP low-level controller (top), the learned policy encounters this failure mode.
With the QP controller (bottom), a hand--table clearance constraint prevents the collision without retraining the task-space policy.
  }
  \label{fig:qp_failure}
\end{figure}

\noindent\textbf{Controller formulation.}
We write superscripts for the frame in which a quantity is expressed and subscripts for the associated body or point. $\mathcal{B}$ and $\mathcal{P}$ denote the robot base and palm frames, and $\mathcal{F}$ denotes the set of fingertips.
The policy outputs
\begin{equation}
    a =
    \left(
        \xi^{\mathcal{B}}_{P},
        \{v^{\mathcal{P}}_{i}\}_{i\in\mathcal{F}}
    \right),
\end{equation}
where $\xi^{\mathcal{B}}_{P}\in\mathbb{R}^{6}$ is the desired palm twist
expressed in the robot base frame and
$v^{\mathcal{P}}_{i}\in\mathbb{R}^{3}$ is the desired linear velocity of
fingertip $i$ relative to the palm. The resulting action space has dimension
$6+3|\mathcal{F}|$.

The low-level QP computes the joint velocity that best realizes these
task-space commands:
\begin{align}
    \dot{q}_{\mathrm{des}}
    =
    \argmin_{\dot{q}}\;&
    \left\|
        \xi^{\mathcal{B}}_{P}
        -
        J^{\mathcal{B}}_{P}(q)\dot{q}
    \right\|_2^2
    +
    \sum_{i\in\mathcal{F}}
    \left\|
        v^{\mathcal{P}}_{i}
        -
        J^{\mathcal{P}}_{i}(q)\dot{q}
    \right\|_2^2
    +
    \lambda\|\dot{q}\|_2^2
    \label{eq:qp_formulation}
    \\
    \text{s.t.}\quad&
    \varphi_c(q)
    +
    \nabla \varphi_c(q)^{\top}\dot{q}\Delta t
    \geq
    \varepsilon_{\mathrm{col}},
    \qquad c\in\mathcal{C},
    \notag\\
    &
    q_{\min}
    \leq
    q+\dot{q}\Delta t
    \leq
    q_{\max},
    \qquad
    \dot{q}_{\min}
    \leq
    \dot{q}
    \leq
    \dot{q}_{\max}.
    \notag
\end{align}
Here, $J^{\mathcal{B}}_{P}$ maps joint velocities to the palm twist, while
$J^{\mathcal{P}}_{i}$ maps them to the palm-relative velocity of fingertip
$i$. Because the fingertip velocities are defined relative to the palm, the arm columns of $J^{\mathcal{P}}_{i}$ vanish, so the corresponding objective terms control only the finger joints. The constraints enforce collision clearance, joint-position limits, and joint-velocity limits over the control interval $\Delta t$. Here, $\varphi_c$ denotes the signed distance for a monitored collision pair $c\in\mathcal{C}$ and $\varepsilon_{\mathrm{col}}$ is the required clearance.

\noindent\textbf{Deployment-time constraint adaptation.}
The separation between task-space policy outputs and constrained joint-space execution allows us to modify the feasible joint motion at deployment without changing the learned policy. In particular, we can tighten $\dot{q}_{\min}$ and $\dot{q}_{\max}$ to reduce execution speed on the physical robot, while preserving the task-space behavior learned in simulation.
We can also introduce additional collision constraints that were not required during simulation training.

For our real-world setup, we add a hand--table clearance constraint
\begin{equation}
    \varphi_{\mathrm{table}}(q)
    +
    \nabla\varphi_{\mathrm{table}}(q)^{\top}\dot{q}\Delta t
    \geq
    \varepsilon_{\mathrm{table}},
    \label{eq:table_constraint}
\end{equation}
where $\varphi_{\mathrm{table}}$ measures the minimum signed distance between
the hand and the table and $\varepsilon_{\mathrm{table}}$ specifies the
required clearance. This constraint modifies only the QP at deployment; the
RL policy remains unchanged.

Figure~\ref{fig:qp_failure} illustrates why this separation is useful in our physical system. Without the hand--table clearance constraint, the learned policy can drive the Wuji 1 hand coupler into the table, which we observed to disengage the coupler during real-world execution. Adding the clearance constraint prevents this failure mode without additional policy training. Thus, the QP provides a mechanism for incorporating deployment-specific hardware and safety requirements while preserving the task-space policy trained in simulation.

\subsection{Simulation Training Details}
\label{app:training_details}

\noindent\textbf{Simulation setup.}
We train all policies in Isaac Sim with Isaac Lab using the GPU PhysX pipeline. The simulator runs at $120$\,Hz, and the policy acts every $12$ simulation steps, resulting in a control frequency of $10$\,Hz. Each episode lasts $16$\,s, corresponding to $160$ policy steps. At every policy step, we solve the QP in Sec.~\ref{app:QP_controller} in batch across all parallel environments to obtain the desired joint velocities. The four evaluated embodiments have $22$--$28$ actuated joints and four or five fingertips. Accordingly, the task-space action dimension is $18$ for four-finger hands and $21$ for five-finger hands.

\noindent\textbf{Policy optimization.}
We optimize the policy with PPO~\citep{ppo} under an asymmetric actor--critic. The actor receives only observations available at deployment and uses a single-layer LSTM with layer normalization followed by a four-layer MLP. The critic is memoryless and receives privileged simulation state. We use the same optimization and network configuration across all tasks and embodiments (Table~\ref{tab:ppo}).

\begin{table}[!tbp]
\centering
\caption{
\textbf{Policy optimization configuration.}
We use the same configuration for all tasks and embodiments.
The learning rate follows an adaptive KL schedule initialized at
$10^{-4}$ and clipped to $[10^{-5},10^{-2}]$.
}
\label{tab:ppo}
\small
\setlength{\tabcolsep}{3.5pt}
\begin{tabular}{llll}
\toprule
\multicolumn{2}{l}{\textbf{PPO}} &
\multicolumn{2}{l}{\textbf{Network}} \\
\midrule
Parallel environments  & $49{,}152$
    & Actor recurrence     & LSTM, 1 layer, width $1024$ \\
Rollout horizon        & $32$ steps ($3.2$\,s)
    & Actor recurrent norm & Layer norm \\
Learning epochs        & $5$
    & Actor MLP            & $[1024,1024,512,512]$ \\
Mini-batches           & $4$
    & Critic               & MLP $[1024,1024,512,512]$ \\
Clip range             & $0.2$
    & Activation           & ELU \\
Discount $\gamma$      & $0.99$
    & Action distribution  & Gaussian, learned $\log\sigma$ \\
GAE $\lambda$          & $0.95$
    & Initial $\sigma$     & $1.0$ \\
Entropy coefficient    & $0.006$
    & Observation history  & $1$ \\
Value-loss coefficient & $1.0$ (clipped)
    &                      & \\
Optimizer              & Adam
    &                      & \\
Initial learning rate  & $1\times10^{-4}$
    &                      & \\
Schedule               & Adaptive, $\mathrm{KL}^{*}=0.01$
    &                      & \\
Gradient-norm clip     & $1.0$
    &                      & \\
Reset states per stage & 10{,}000 \\
\bottomrule
\end{tabular}
\end{table}

\noindent\textbf{Domain randomization.}
During simulation training, we vary physical properties and deliberately add noise to the actor's observations to improve robustness to real-world differences. Also we introduce observation and action delays to deal with sim2real gap. Table~\ref{tab:dr} summarizes these variations.

\begin{table}[!tpb]
\centering
\caption{\textbf{Domain randomization.}
Multipliers are relative to the original simulation values.
Ranges are sampled uniformly unless otherwise stated.}
\label{tab:dr}
\small
\renewcommand{\arraystretch}{1.15}
\begin{tabularx}{\linewidth}{
    @{}l >{\raggedright\arraybackslash}X@{}
}
\toprule
\textbf{Property} & \textbf{How it is varied} \\
\midrule
\multicolumn{2}{@{}l}{\textit{Physical properties}} \\
\addlinespace[2pt]

Object mass
  & Between the minimum and maximum mass estimated for each object \\

Object static friction
  & Up to 10\% above or below the estimated friction \\

Object dynamic friction
  & Up to 10\% above or below the estimated friction \\

Table friction
  & Static: 0.3--0.6; dynamic: 0.2--0.5 \\

Table mass
  & 0.5--1.5 times the original mass \\

Hand stiffness and damping
  & 0.8--1.25 times the original values \\

\midrule
\multicolumn{2}{@{}l}{\textit{Errors in actor observations}} \\
\addlinespace[2pt]

object position
  & Gaussian error: 1 cm standard deviation per axis;
    capped at 3 cm per axis \\

static object position
  & Gaussian error: 0.5 cm standard deviation per axis;
    capped at 1.5 cm per axis \\

Object orientation
  & Rotate about a uniformly sampled direction
    by an angle between $-5^\circ$ and $+5^\circ$ \\

Perceived object size
  & Up to 10\% larger or smaller, equally in all directions \\

\midrule
\multicolumn{2}{@{}l}{\textit{Delays}} \\
\addlinespace[2pt]

Object-pose observation & 0--150 ms \\
Robot-state observation & 0--50 ms \\
Action execution        & 0--50 ms \\

\bottomrule
\end{tabularx}

\vspace{4pt}
\end{table}

\newpage

\section{Real-World Perception}
\label{app:real}

The policy observes object poses rather than images (Sec.~\ref{sec:rl}), so at deployment we estimate the 6D pose of each task-relevant object in real time. Two RGB-D cameras (Intel RealSense D435i) observe the workspace from complementary viewpoints, and FoundationPose++~\citep{FoundationPose-plus-plus} tracks each object independently in each camera, using the same mesh as in simulation. Because the hand frequently occludes the object during manipulation, a single view is often unreliable, and we fuse the two estimates as follows.

\paragraph{Multi-view pose fusion.}
\label{app:pose_fusion}
We fuse the FoundationPose++ estimates $T_1, T_2$ from the two cameras according to how well each explains its own depth image. For camera $c$, we render the object mesh at $T_c$ and compute the fraction of its projected silhouette $\Omega_c$, including the part outside the image, whose sensor depth is valid and within $\tau$ of the rendered depth:
\begin{equation}
w_c = \frac{1}{|\Omega_c|} \sum_{u \in \Omega_c} \mathbb{I}\big[\, |D_c(u) - \hat{D}_c(u)| < \tau \,\big].
\end{equation}
Occluded or truncated views thus receive low weights. Because the attainable value differs between objects and cameras even without occlusion (a white sheet, for instance, returns no depth over much of its surface), each weight is normalized by its unoccluded baseline $w_c^0$, the median over the first 15 frames: $\bar w_c = \min\!\big(w_c / \max(w_c^0, 0.1),\, 1\big)$. The fused pose interpolates $T_1$ and $T_2$ with $\alpha = \bar w_2^{k} / (\bar w_1^{k} + \bar w_2^{k})$, linearly in translation and by slerp in rotation. A camera is excluded when $\bar w_c < w_{\min}$ and readmitted only once $\bar w_c > w_{\mathrm{re}}$, and a pose that deviates from the higher-weighted one by more than 3\,cm or $20^\circ$ is left out of the blend, so that a lost track or a symmetric flip is never averaged. If both cameras are excluded, the pose is extrapolated with a constant-velocity model for up to 0.5\,s and then held. The fused pose also serves as the next tracking prior of both cameras, which re-initializes a camera after an occlusion. We use $\tau = 1.5$\,cm, $k = 3$, $w_{\min} = 0.25$, and $w_{\mathrm{re}} = 0.35$.

\section{Additional Experimental Results}

\subsection{Generalization to Unseen Initial Object Positions}
\label{app:initial_generalization}
\begin{figure}[h]
    \centering
    \includegraphics[width=0.9\linewidth]
    {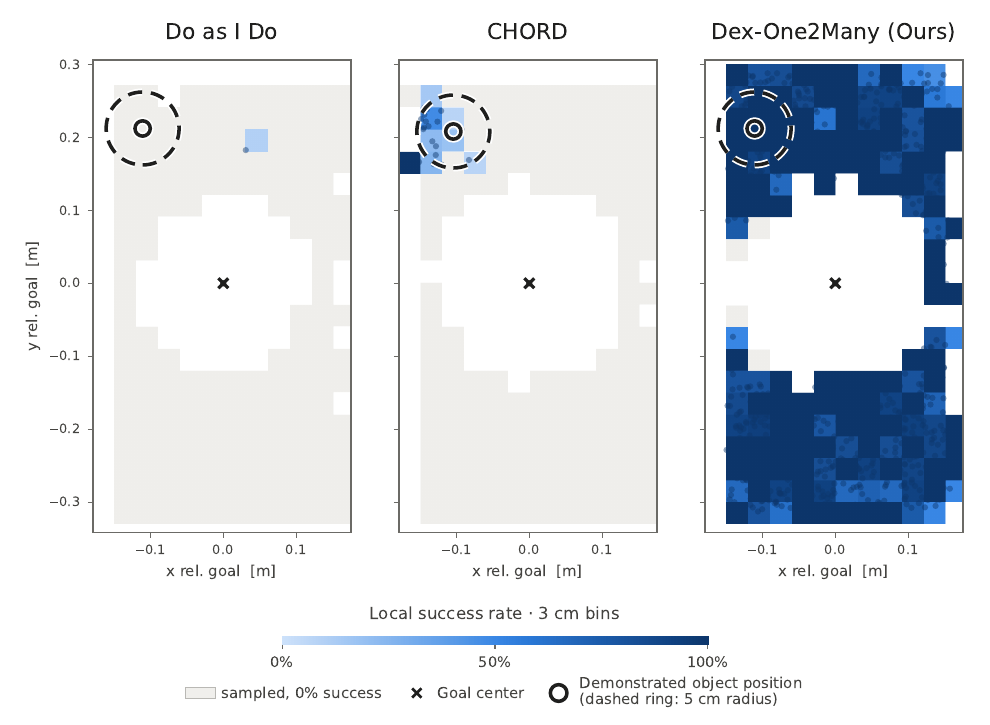}
    \caption{
    \textbf{Generalization to unseen initial object positions on the Doll task.}
    With the goal fixed, \algoname succeeds over a substantially broader range
    of initial positions than the baselines.
    }
    \label{fig:initial_landscape_doll}
\end{figure}

\begin{figure}[p]
    \centering
    \includegraphics[width=0.9\linewidth]
    {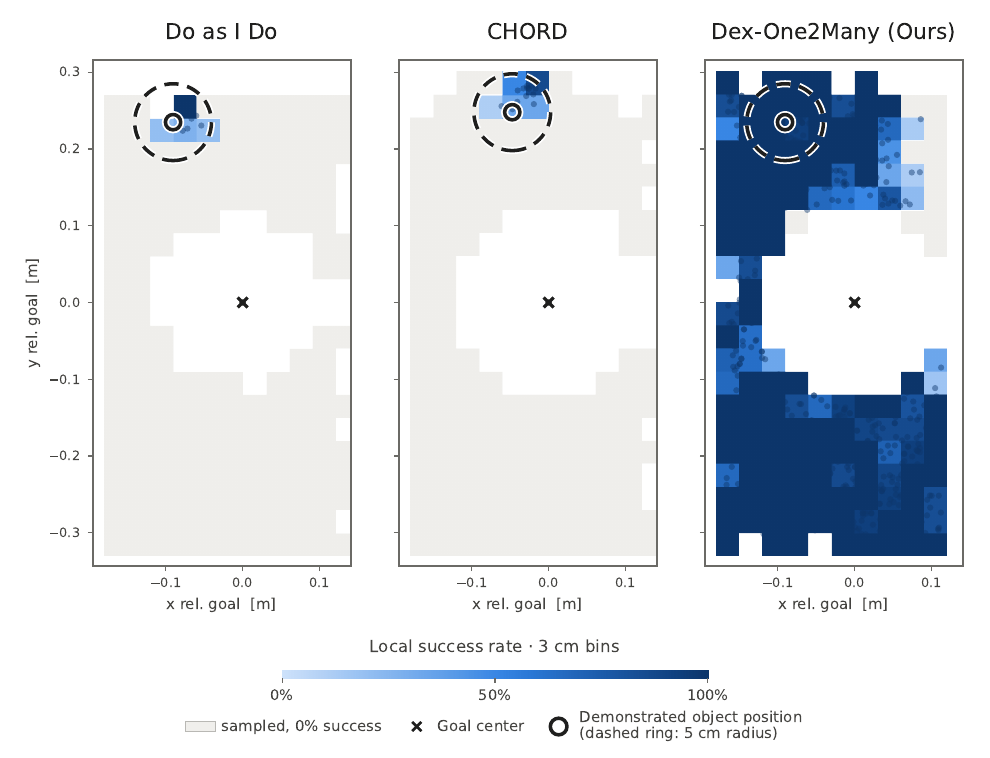}
    \caption{
    \textbf{Generalization to unseen initial object positions on the Can task.}
    With the goal fixed, \algoname succeeds over a substantially broader range
    of initial positions than the baselines.
    }
    \label{fig:initial_landscape_can}
\end{figure}

\begin{figure}[p]
    \centering
    \includegraphics[width=0.9\linewidth]
    {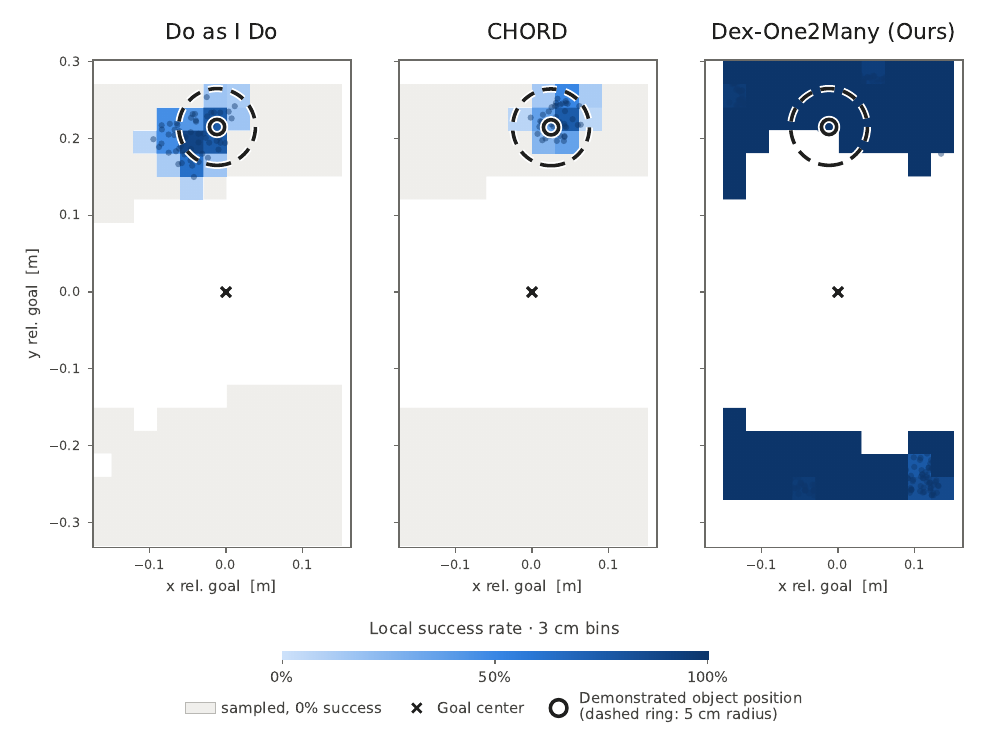}
    \caption{
    \textbf{Generalization to unseen initial object positions on the Stamp task.}
    With the goal fixed, \algoname succeeds over a substantially broader range
    of initial positions than the baselines.
    }
    \label{fig:initial_landscape_stamp}
\end{figure}

\begin{figure}[p]
    \centering
    \includegraphics[width=0.9\linewidth]
    {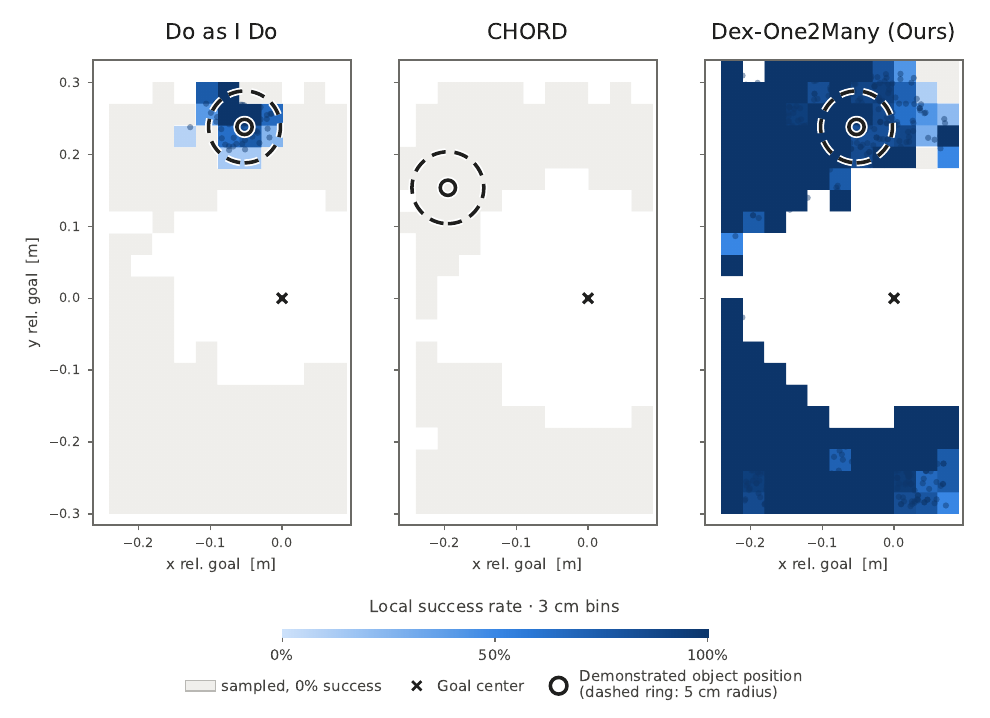}
    \caption{
    \textbf{Generalization to unseen initial object positions on the Hammer task.}
    With the goal fixed, \algoname succeeds over a substantially broader range
    of initial positions than the baselines.
    }
    \label{fig:initial_landscape_hammer}
\end{figure}

\begin{figure}[p]
    \centering
    \includegraphics[width=0.9\linewidth]
    {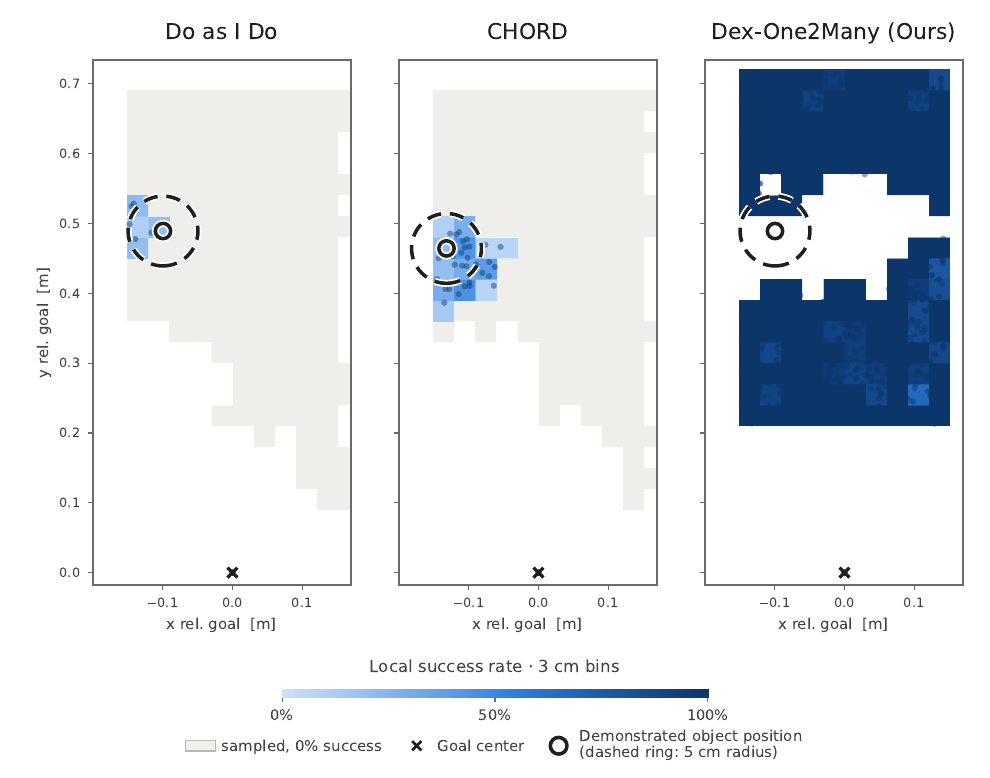}
    \caption{
    \textbf{Generalization to unseen initial object positions on the Sweep task.}
    With the goal fixed, \algoname succeeds over a substantially broader range
    of initial positions than the baselines.
    }
    \label{fig:initial_landscape_sweep}
\end{figure}

To further analyze spatial generalization beyond the human video, we fix the goal configuration and vary the initial object position across the feasible workspace. Figures~\ref{fig:initial_landscape_doll}--\ref{fig:initial_landscape_sweep}
visualize initial-position generalization across the five evaluation tasks.

\FloatBarrier

\FloatBarrier

\clearpage

\subsection{Diverse Grasp Strategies Across Embodiments and Configurations}
\label{app:additional_diverse_grasps}

The task specification extracted from the human video does not prescribe a specific grasp configuration. Instead, our reset generation pipeline exposes the policy to diverse multi-finger grasps drawn from the 33-type grasp taxonomy, allowing the grasp strategy to adapt to the robot embodiment and the current object configuration.

\begin{figure}[!htbp]
    \centering
    \includegraphics[width=\linewidth]{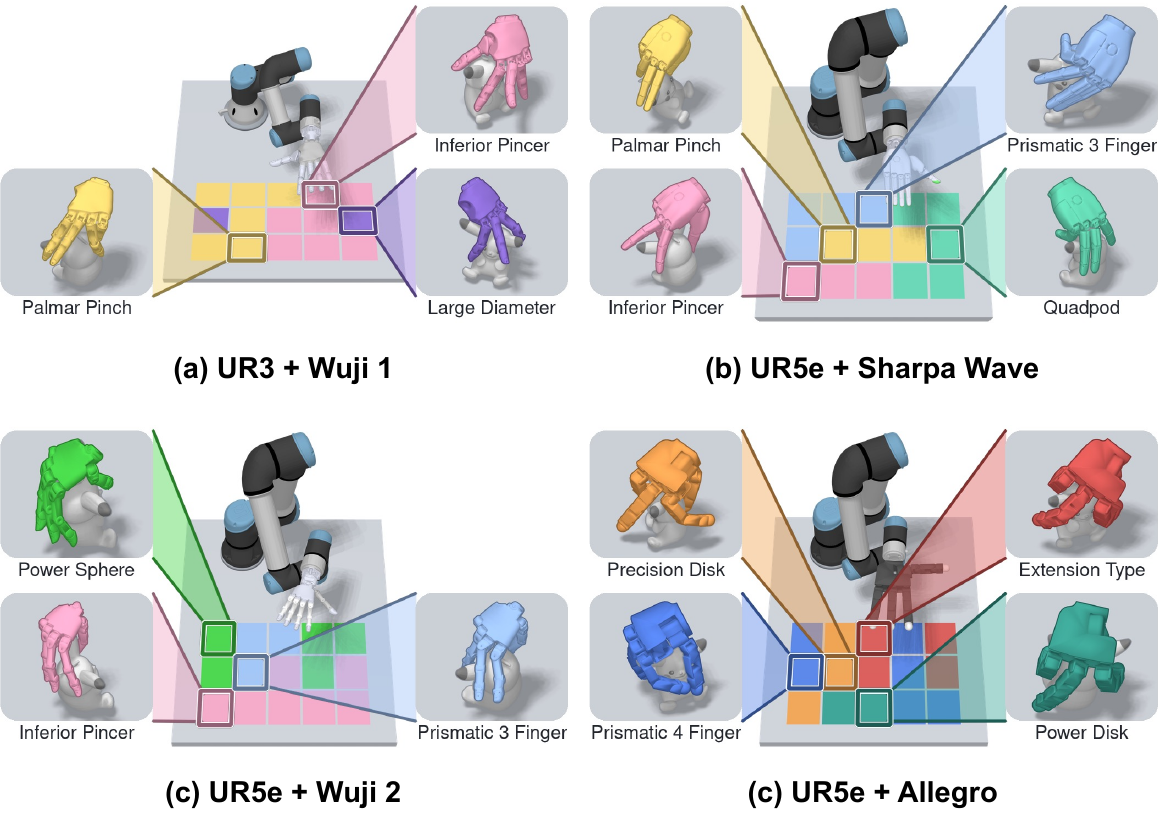}
    \caption{
    \textbf{Diverse grasp strategies for the Doll task.}
    Learned grasp types vary across dexterous embodiments and object configurations.
    }
    \label{fig:diverse_grasp_doll}
\end{figure}

\begin{figure}[p]
    \centering
    \includegraphics[width=\linewidth]{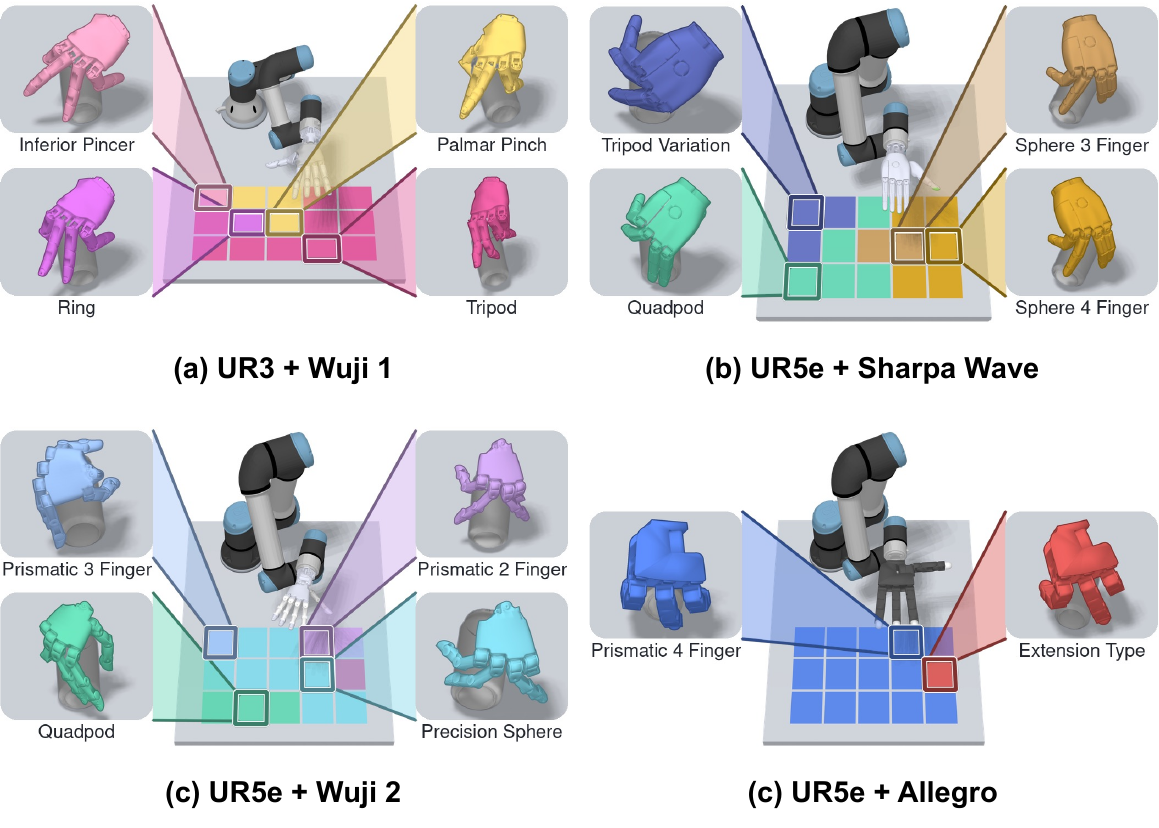}
    \caption{
    \textbf{Diverse grasp strategies for the Can task.}
    Learned grasp types vary across dexterous embodiments and object configurations.
    }
    \label{fig:diverse_grasp_can}
\end{figure}

\begin{figure}[p]
    \centering
    \includegraphics[width=\linewidth]{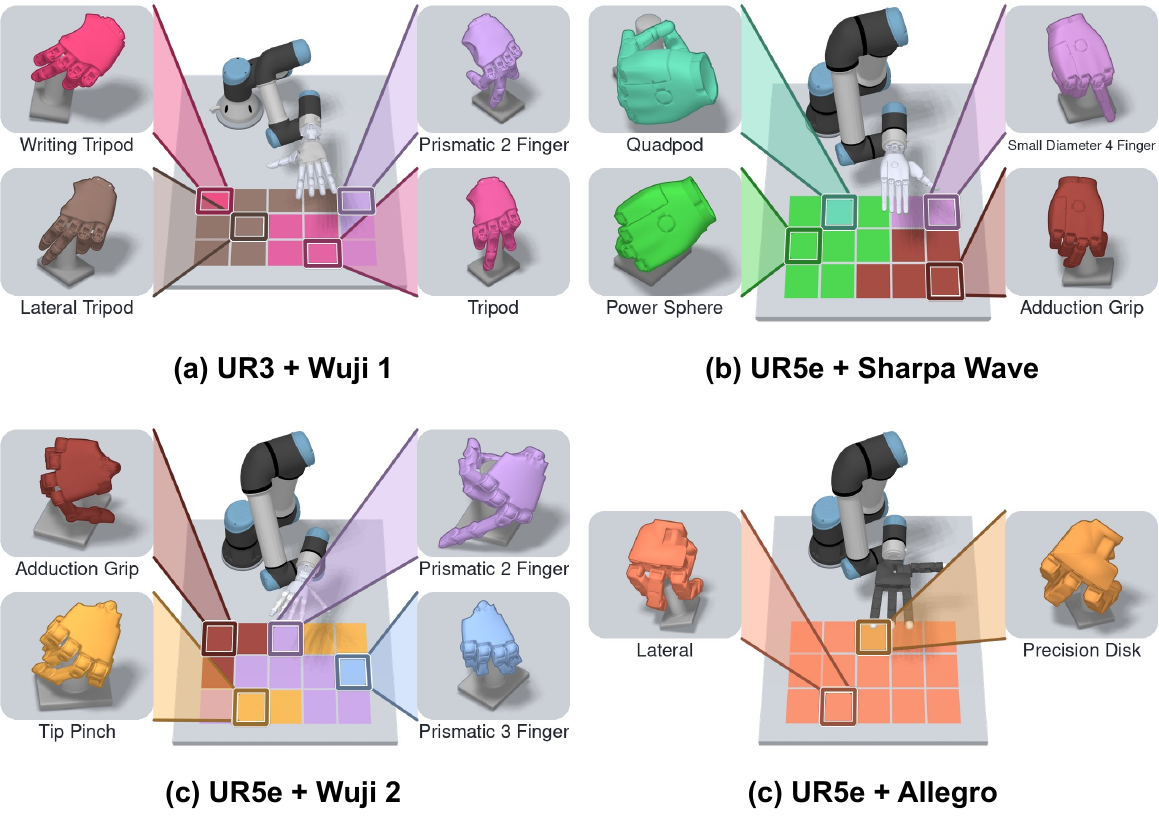}
    \caption{
    \textbf{Diverse grasp strategies for the Stamp task.}
    Learned grasp types vary across dexterous embodiments and object configurations.
    }
    \label{fig:diverse_grasp_stamp}
\end{figure}

Figures~\ref{fig:diverse_grasp_doll}--\ref{fig:diverse_grasp_stamp} show representative grasp strategies learned for the Doll, Can, and Stamp tasks. Different dexterous embodiments realize different grasp types for the same task, reflecting their distinct hand morphologies and kinematics. Moreover, even within the same embodiment, the learned grasp can change as the object configuration varies. These qualitative results illustrate that \algoname does not constrain the robot to a fixed grasp inherited from the human demonstration, but instead allows embodiment- and configuration-specific grasp strategies to emerge during policy learning.

\FloatBarrier

\FloatBarrier

\subsection{Simulation Rollouts Across Diverse Dexterous Embodiments}
\label{app:additional_cross_embodiment}
\begin{figure}[ht]
    \centering
    \includegraphics[width=\linewidth,
                 height=\dimexpr\textheight-6\baselineskip\relax,
                 keepaspectratio]
    {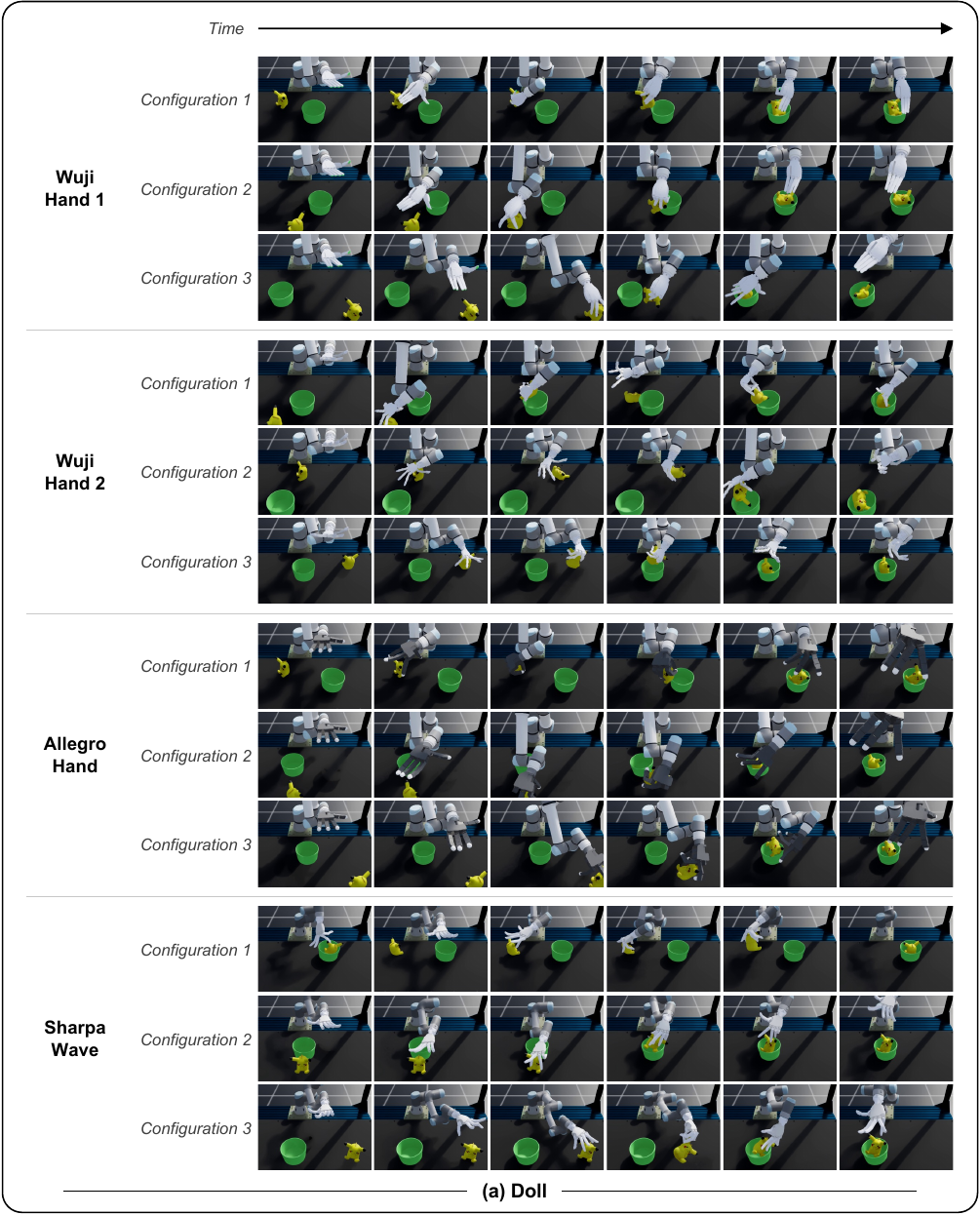}
    \caption{\textbf{Successful simulation rollouts across diverse dexterous embodiments for the Doll task.} For each of four dexterous embodiments, we show three successful rollouts from distinct task configurations. Each row corresponds to one configuration, and columns show the execution over time from left to right.}
    \label{fig:appendix_simulation_figure_doll}
\end{figure}

\begin{figure}[p]
    \centering
    \includegraphics[width=\linewidth,
                 height=\dimexpr\textheight-6\baselineskip\relax,
                 keepaspectratio]
    {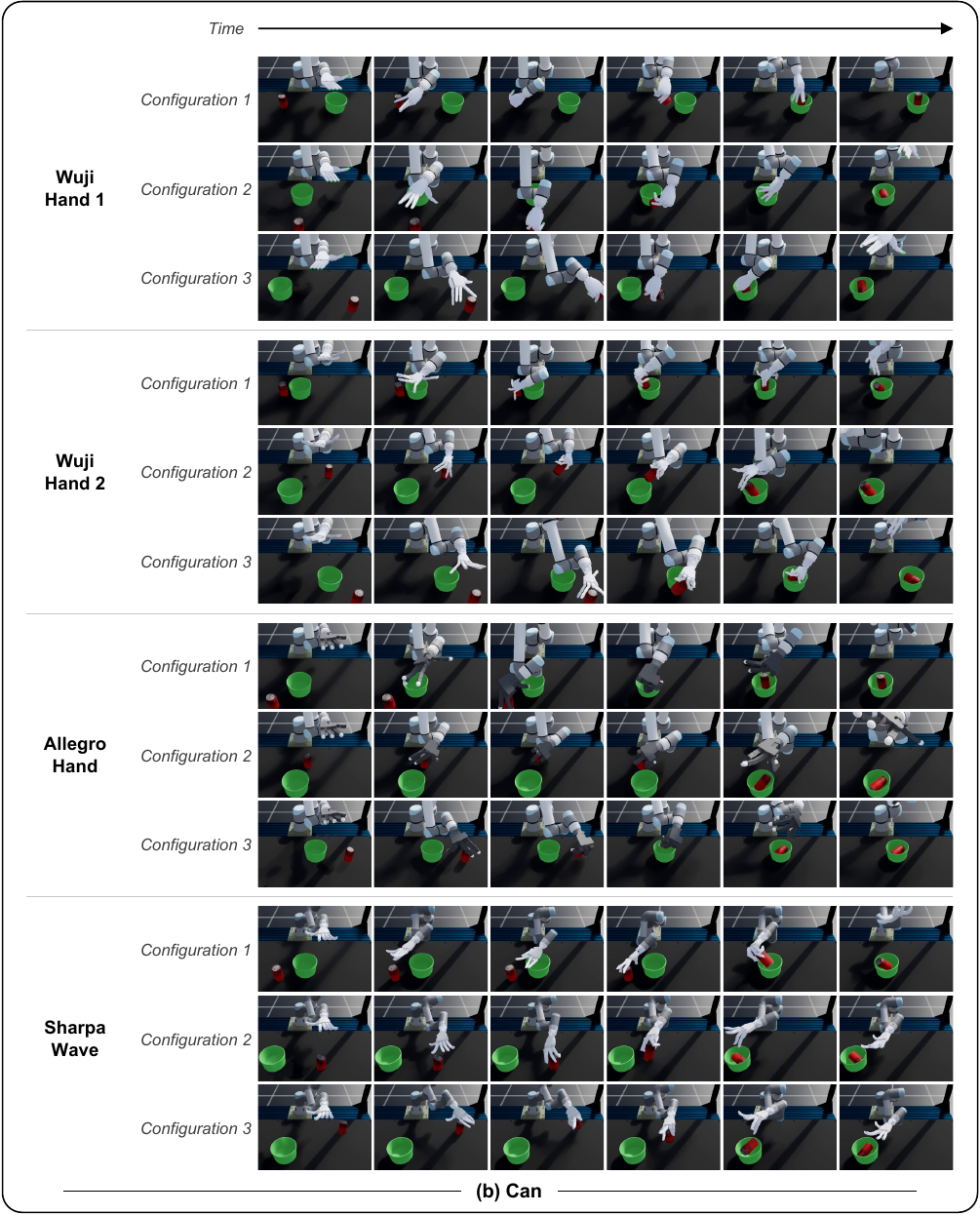}
    \caption{\textbf{Successful simulation rollouts across diverse dexterous embodiments for the Can task.}
For each of four dexterous embodiments, we show three successful rollouts from distinct task configurations.
Each row corresponds to one configuration, and columns show the execution over time from left to right.}
    \label{fig:appendix_simulation_figure_can}
\end{figure}

\begin{figure}[p]
    \centering
    \includegraphics[width=\linewidth,
                 height=\dimexpr\textheight-6\baselineskip\relax,
                 keepaspectratio]
    {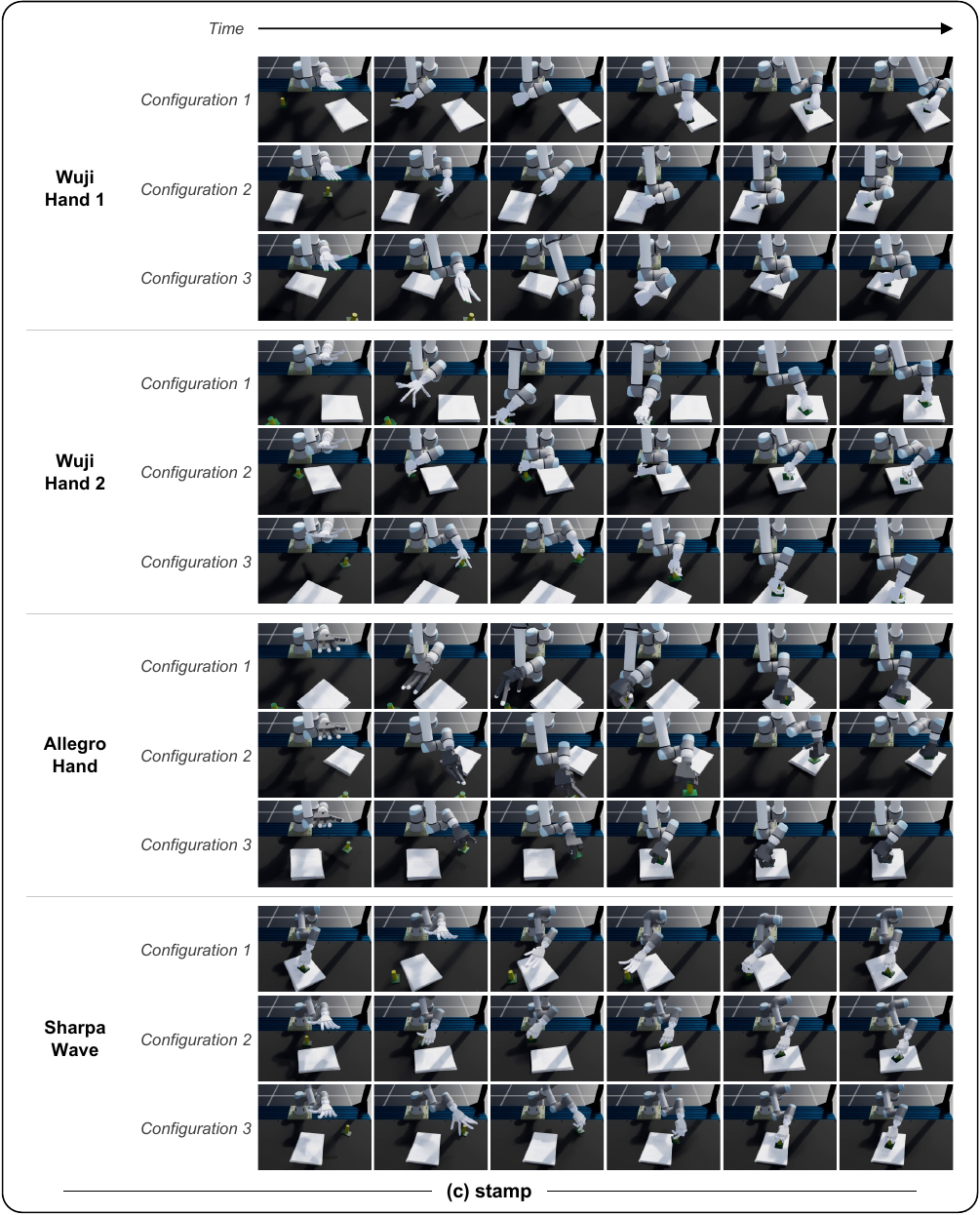}
    \caption{
\textbf{Successful simulation rollouts across diverse dexterous embodiments for the Stamp task.}
For each of four dexterous embodiments, we show three successful rollouts from distinct task configurations.
Each row corresponds to one configuration, and columns show the execution over time from left to right.
}
    \label{fig:appendix_simulation_figure_stamp}
\end{figure}

\begin{figure}[p]
    \centering
    \includegraphics[width=\linewidth,
                 height=\dimexpr\textheight-6\baselineskip\relax,
                 keepaspectratio]
    {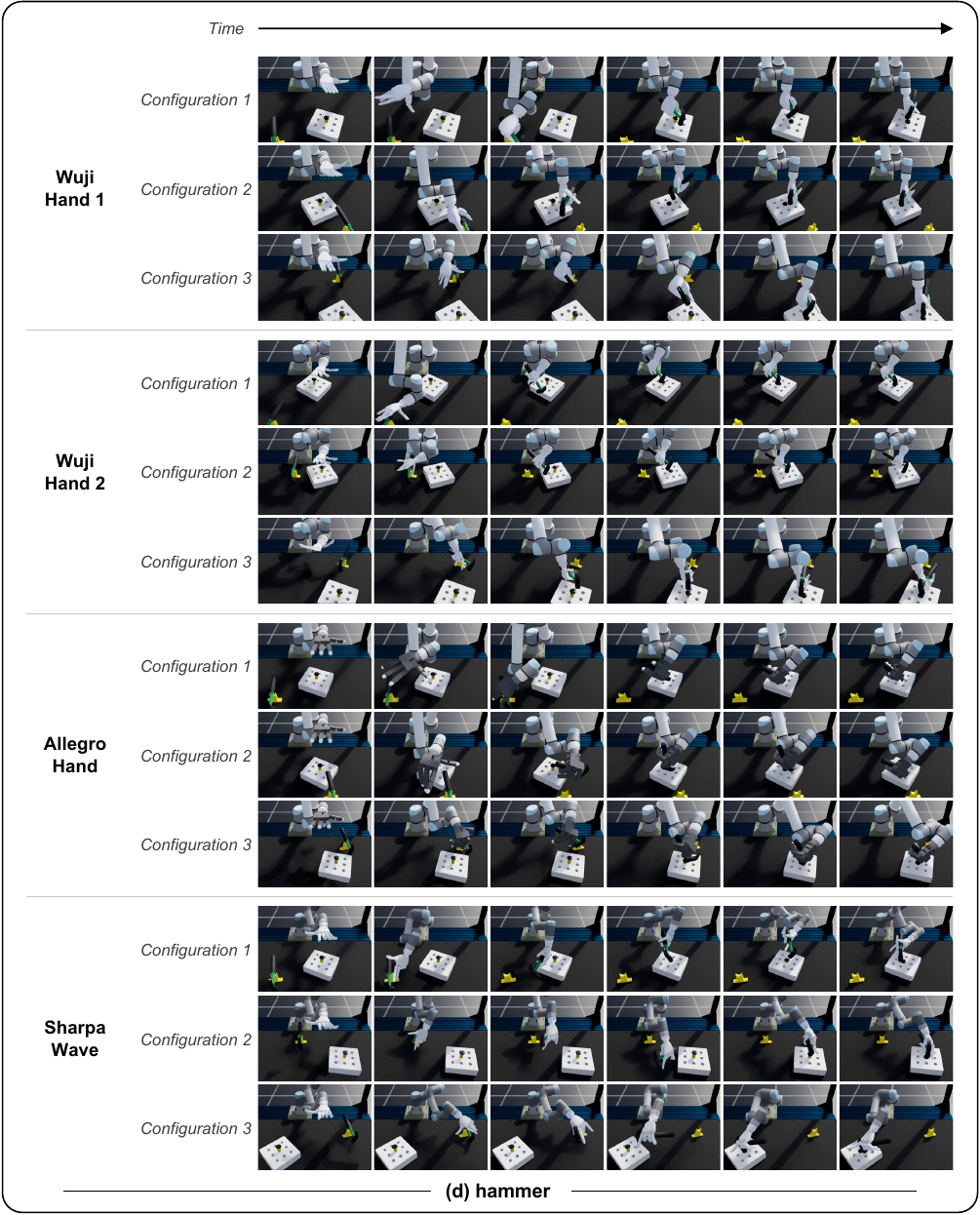}
    \caption{
\textbf{Successful simulation rollouts across diverse dexterous embodiments for the Hammer task.}
For each of four dexterous embodiments, we show three successful rollouts from distinct task configurations.
Each row corresponds to one configuration, and columns show the execution over time from left to right.
}
    \label{fig:appendix_simulation_figure_hammer}
\end{figure}

\begin{figure}[p]
    \centering
    \includegraphics[width=\linewidth,
                 height=\dimexpr\textheight-6\baselineskip\relax,
                 keepaspectratio]
    {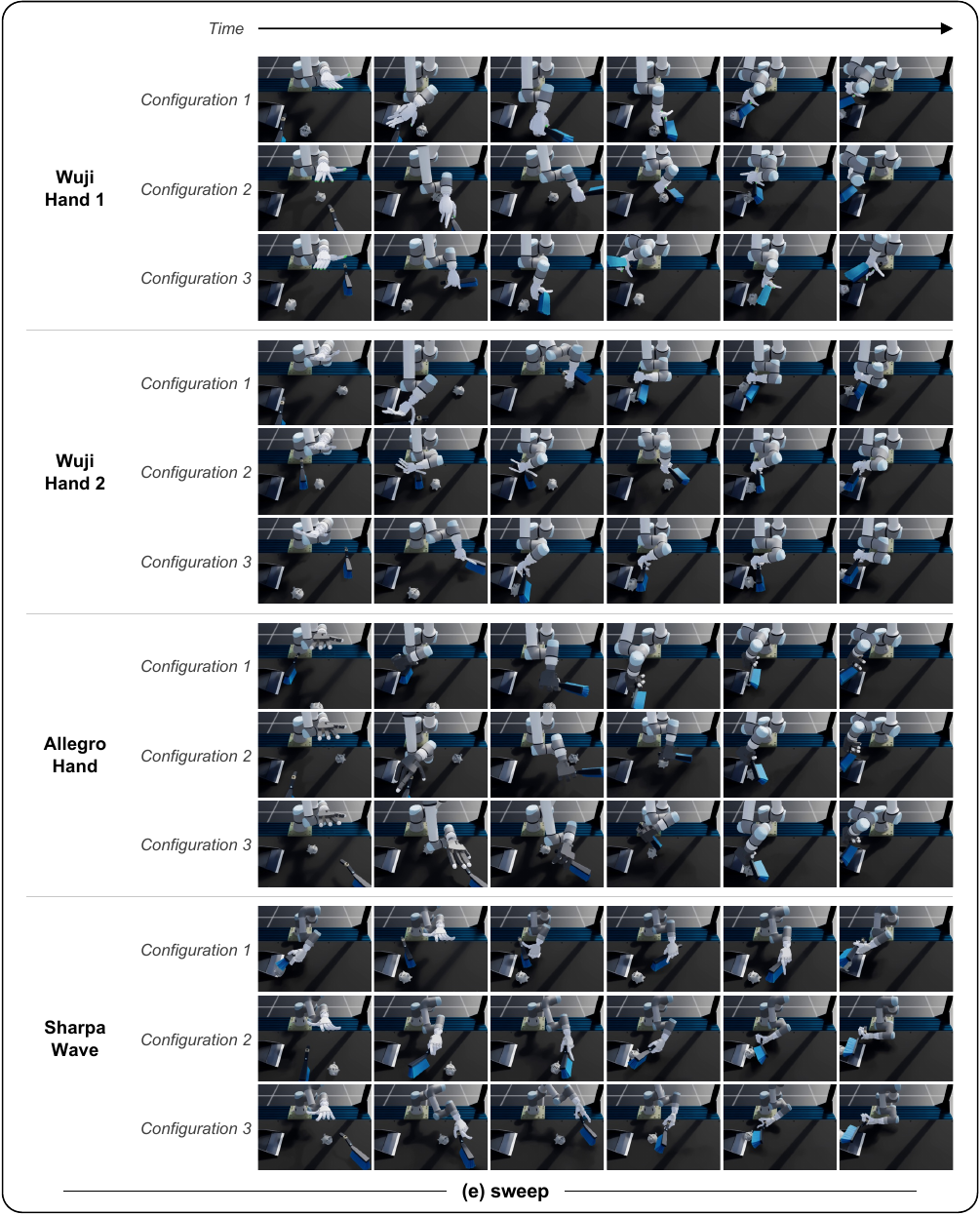}
    \caption{
\textbf{Successful simulation rollouts across diverse dexterous embodiments for the Sweep task.}
For each of four dexterous embodiments, we show three successful rollouts from distinct task configurations.
Each row corresponds to one configuration, and columns show the execution over time from left to right.
}
    \label{fig:appendix_simulation_figure_sweep}
\end{figure}

To complement the quantitative evaluation in Table~\ref{tab:cross-embodiment}, we visualize successful simulation rollouts across four dexterous embodiments for all five tasks.
For each embodiment and task, we show three rollouts initialized from distinct task configurations.
Each row corresponds to one configuration, and the columns show the execution over time from left to right. 
Figure~\ref{fig:appendix_simulation_figure_doll}--\ref{fig:appendix_simulation_figure_sweep} show that separately trained policies execute the same task configuration across embodiments with different kinematics and under diverse configurations.

\clearpage

\clearpage

\clearpage

\clearpage

\end{document}